\documentclass[aoas]{imsart}

\RequirePackage{amsthm,amsmath,amsfonts,amssymb}
\RequirePackage[authoryear]{natbib}
\RequirePackage[colorlinks,citecolor=blue,urlcolor=blue]{hyperref}
\RequirePackage{graphicx}

\usepackage{mathrsfs}
\usepackage{bm}
\usepackage{algorithm}
\usepackage{algorithmic}
\usepackage{multirow}
\usepackage{booktabs}
\usepackage[most]{tcolorbox}
\usepackage{prodint}

\graphicspath{{figures/}}

\startlocaldefs
\theoremstyle{plain}

\newtheorem{theorem}{Theorem}[section]

\theoremstyle{definition}
\newtheorem{definition}[theorem]{Definition}

\newtheorem{remark}[theorem]{Remark}
\def\AUC{\mathrm{AUC}}
\def\bbeta{\boldsymbol{\beta}}

\newcommand{\hhat}{\widehat{\lambda}}
\newcommand{\Hhat}{\widehat{\Lambda}}
\newcommand{\haz}{\lambda}
\newcommand{\Haz}{\Lambda}

\renewcommand{\tt}{\mathcal T}

\def\x{{\bf x}}
\def\X{{\bf X}}

\def\Z{{\bf Z}}

\def\iid{\stackrel{\text{iid}}{\sim}}

\def\wp1{\stackrel{\text{with probability one}}\rightarrow}
\def\E{\mathbb{E}}
\def\PP{\mathbb{P}}
\def\RR{\mathbb{R}}

\def\RSF{\mathscr}

\def\oo{{\RSF O}}

\def\ss{{\RSF S}}
\def\tt{{\RSF T}}

\def\a{\alpha}\def\d{\delta}\def\D{\Delta}

\newcommand{\FancyPiScale}[1]{\mathord{\mathpalette\FancyPiScaleAux{#1}}}
\newcommand{\FancyPiScaleAux}[2]{%
  \begingroup
  \setbox0=\hbox{\scalebox{#2}{$#1\Prodi$}}%
  \raise\dp0\box0%
  \endgroup
}
\def\Pi{\FancyPiScale{.480}}
\def\Pismall{\FancyPiScale{.440}}
\def\bct{\begin{center}}
\def\ect{\end{center}}
\def\Array{\begin{eqnarray*}}
\def\EndArray{\end{eqnarray*}}
\def\Enumerate{\begin{enumerate}}
\def\EndEnumerate{\end{enumerate}}
\def\Eq{\begin{equation}}
\def\EndEq{\end{equation}}
\def\EqArray{\begin{eqnarray}}
\def\EndEqArray{\end{eqnarray}}
\def\Itemize{\begin{itemize}}
\def\EndItemize{\end{itemize}}

\def\Tabular{\begin{tabular}}
\def\EndTabular{\end{tabular}}
\def\FlushLeft{\begin{flushleft}}
\def\EndFlushLeft{\end{flushleft}}
\def\({\left(}
\def\){\right)}
\def\[{\left[}
\def\]{\right]}

\endlocaldefs

\begin{document}


\begin{frontmatter}
\title{Random Hazard Forests}
\runtitle{Random Hazard Forests}

\begin{aug}
\author[A]{\fnms{Hemant}~\snm{Ishwaran}\ead[label=e1]{hishwaran@miami.edu}\orcid{0000-0003-2758-9647}},
\author[B]{\fnms{Eileen M.}~\snm{Hsich}\ead[label=e2]{eileen.hsich@imail.org}},\\
\author[C]{\fnms{Udaya B.}~\snm{Kogalur}\ead[label=e3]{ubkogalur@gmail.com}}
\and
\author[D]{\fnms{Donald K.~K.}~\snm{Lee}\ead[label=e4]{donald.lee@emory.edu}}

\address[A]{Division of Biostatistics,
Department of Public Health Sciences,
Miller School of Medicine,
University of Miami\printead[presep={,\ }]{e1}}

\address[B]{Heart and Vascular Institute,
Cleveland Clinic; Cardiology Division,
Intermountain Health%
\printead[presep={,\ }]{e2}}

\address[C]{Kogalur \& Company, Inc.\printead[presep={,\ }]{e3}}

\address[D]{%
  Goizueta Business School and Department of Biostatistics \& Bioinformatics, 
  Emory University\printead[presep={,\ }]{e4}}
\end{aug}


\begin{abstract}

Clinical data sources such as electronic health records and wearable
sensors record patient status repeatedly over follow-up, often at
irregular times and on different schedules for different
measurements. These data create opportunities for continuously
updated, individualized risk prediction. Existing approaches, however,
often simplify the temporal structure for model fitting. We
introduce Random Hazard Forests (RHF), a survival
tree ensemble that estimates how a patient's hazard changes in continuous
time as new measurements become available. The method formulates the
estimation problem directly through a nonparametric hazard likelihood
for predictable covariate processes. An efficient working model guides
tree construction, after which flexible time-varying hazards are
estimated for each terminal node. Given any predictable covariate
path, each tree follows the path through its terminal nodes over time
and assembles the corresponding node-level hazards into a trajectory.
Averaging these trajectories across trees yields the pathwise
hazard estimate. Because routing at each time uses only the covariate
state available immediately beforehand, the construction accommodates
internal longitudinal covariates without lookahead. Simulations and an
intensive care application show that the forest accurately estimates
changing risk under irregular and asynchronous covariate updates.

\end{abstract}


\end{frontmatter}


\section{Introduction}

Clinical risk is rarely static.  Yet many established survival methods
predict risk from covariates measured once at study entry, reducing
each subject's history to a single baseline snapshot.  That
simplification was natural when follow-up data were sparse, but it
fits poorly with modern clinical data systems.  In the ICU, a
patient's physiology, laboratory values, and organ support therapies
can shift from hour to hour, and these shifts often signal impending
deterioration.  More broadly, clinical trials collect repeated
assessments, wearable devices stream physiological measurements, and
electronic health records accumulate information over days, months, or
years.  In all of these settings, the question is not only who is high
risk at study entry, but how risk changes as new information arrives.
Analyses that condition only on an initial snapshot can therefore miss
some of the most important changes in patient status.

Using evolving records in event-time analysis, however, poses
challenges.  Measurements arrive irregularly, different
variables update on different schedules, and treatments or exposures
may start, stop, and restart over follow-up.  For prospective
prediction, the most important constraint is temporal order.  A risk
estimate at time $t$ should depend only on information available up to
$t$.  If later measurements enter the calculation, the model may
appear accurate in retrospective evaluation but fail when deployed in
real time~\citep{ramadan2025diagnostic}.

These issues motivate methods that can follow longitudinal covariate
histories while preserving the no-lookahead structure of prospective
estimation.  Tree ensembles are a natural starting point, since they
handle high-dimensional covariates, nonlinearities, and interactions
with little tuning.  Standard tree methods, however, are built around a
single fixed covariate vector for each subject.  Once predictors vary
over time, tree-based learning faces two challenges: how to choose
splits from longitudinal records, and how to assign a time-varying
covariate to a terminal node at each time.  Both must be resolved using
only the information available at that time.  This paper introduces
\emph{Random Hazard Forests} (RHF), a survival tree ensemble for this
setting.  The method works in continuous time and estimates how the hazard varies
with time and the current predictable covariate state, rather than
reducing each subject's history to a baseline snapshot.

RHF represents this relationship through a forest hazard map on
time--covariate space,
$$
(t,\x)\longmapsto\hhat_{\rm RHF}(t,\x).
$$
Given a time $t$ and a covariate state $\x$, the map returns an
estimated hazard by routing $\x$ to a terminal node in each tree and
averaging the node hazards.  A case-specific trajectory
follows from evaluating this map along a supplied predictable covariate
process $\X(\cdot)$.  Each tree follows the path through its terminal
nodes over time, and averaging the tree-specific trajectories gives the
forest hazard trajectory
$$
t\longmapsto\hhat_{\rm RHF}(t,\X(t)).
$$
At each time, this is the estimated instantaneous event rate associated
with the current covariate state.  In prospective monitoring, the
supplied path may consist of vital signs, laboratory values, and
treatments observed up to the present, with the hazard trajectory
updating as new information arrives.  In retrospective or
scenario-based analyses, it may instead be an observed history or a
prespecified hypothetical path, and the trajectory then shows how the
hazard evolves across covariate patterns.

We estimate this map by minimizing an empirical risk derived from a
nonparametric hazard likelihood.  This risk guides both tree splitting
and terminal-node hazard estimation and produces a new type of forest
construction for event analysis with longitudinal predictors.
The next two subsections place our proposed method among survival machine
learning methods and tree-based approaches to dynamic prediction.
Section~\ref{sec:contributions} then summarizes the main contributions
of the paper.

\subsection{Background literature}

Classical survival analysis is dominated by parametric and
semiparametric regression models, of which the Cox proportional hazards
model~\citep{cox1972regression} is the most widely used.  In their
standard forms, these models use baseline covariates and time-constant
coefficients.  The Cox model can be extended to time-updated covariates
through the counting-process formulation of \citet{andersen1982cox}, but
the usual extension still relies on a log-linear predictor with
time-constant coefficients.  Joint modeling of longitudinal measurements
and event times \citep{wulfsohn1997joint,henderson2000joint} provides an
alternative by estimating both processes simultaneously through shared
random effects.  These methods, however, require specification of both
the longitudinal trajectory model and the association mechanism, which
becomes more difficult as the number of markers and their interactions
grows.

A separate line of work uses likelihood-based machine learning for
nonparametric hazard estimation with time-dependent covariates.
Closest to the work here is \citet{lee2021boosted}, who developed a
fully nonparametric hazard likelihood for predictable covariate
processes and used its functional gradient to construct boosted hazard
estimators.  That work provides a general likelihood foundation for
nonparametric hazard learning with time-dependent covariates under the
Aalen multiplicative intensity framework that includes right-censoring
as a special case.  More recently, \citet{hu2023conditional} used
neural networks with a full likelihood and an expanded time-varying
covariate data structure to estimate a nonparametric conditional
hazard and survival function.  For the conditional survival target,
that method is developed under an external-covariate, right-censoring
formulation.  The method proposed here shares the nonparametric
hazard-likelihood perspective with these lines of work, but brings it
into the survival tree and random forest setting.

Deep learning has also generated a large literature on survival
prediction.  A number of neural survival methods recast survival
analysis as a surrogate classification task
\citep{ebell1993artificial,tu1993use,liestbl1994survival,zhao2020deep}
or build directly on the Cox model
\citep{faraggi1995neural,luck2017deep,yousefi2017predicting,ching2018cox,katzman2018deepsurv,wang2019extreme}.
Others adopt a discrete-time parameterization
\citep{gensheimer2019scalable,lee2018deephit,jarrett2018match,ren2019deep},
which is computationally convenient and accommodates updated inputs
but replaces the continuous-time problem with prediction on a grid
that must be prespecified.  As the grid is refined and the number of
predictors grows, computation becomes more demanding and performance
may degrade in real-time settings~\citep{nowroozilarki2021real}.

\subsection{Random survival forests and dynamic prediction}

Among machine learning approaches, tree ensembles offer much of the
flexibility of deep models with less architectural complexity and
easier interpretability.  Random Survival Forests
(RSF)~\citep{ishwaran2008rsf} are a widely used example, providing a
nonparametric approach to survival prediction by growing ensembles of
trees that capture nonlinear patterns and interactions with little
tuning.  RSF was designed for baseline
covariates, however.  With time-varying predictors, each subject
contributes a trajectory rather than a single value, and the key
question is how a tree should route an individual at a particular time
using only information observed up to that time.

Survival-tree methods with time-dependent covariates predate modern
forest implementations.  \citet{bacchetti1995survival} introduced a
pseudo-subject construction that allows a subject's risk-set
contribution to move across tree nodes as covariates change over time.
This construction is a precursor to our method, which uses a related
counting-process representation to evaluate the hazard likelihood.

A growing literature has adapted RSF and related forest methods for
dynamic prediction.  One line of work uses landmarking
\citep{van2007dynamic}, fitting a separate RSF at each landmark time
to subjects still at risk~\citep{pickett2021random}.  This yields
updated predictions, but it produces a sequence of landmark-specific
models rather than a single forest over the full follow-up.  Another
line of work handles time-updated covariates by first partitioning
follow-up into prespecified person-period intervals and then fitting
interval-level event models.  One such method is RF-SLAM, which
rewrites each subject's history into counting process information
units (CPIUs), that is, person-period records for prespecified time
bins, and grows trees using a Poisson log-likelihood split rule for a
piecewise-constant hazard model~\citep{wongvibulsin2020clinical}.
Although event times may occur within an interval and interval length
enters as exposure, the forest is still built around interval-specific
hazards rather than exact-event-time risk sets.  Other approaches
summarize longitudinal histories through derived features, for example
via functional principal components, and then apply a tree ensemble to
those summaries \citep{lin2021functional,jiang2021functional}.
Software such as \texttt{DynForest} similarly transforms
time-dependent predictors into node-specific time-fixed features using
mixed models before splitting~\citep{devaux2023dynforest}.
\citet{yao2022ensemble} extend conditional inference forests to
time-varying covariates in a framework targeting the conditional
survival function.

\subsection{Contributions}
\label{sec:contributions}

The methods above show that forests can use evolving covariate
information, but they do so by altering the problem.  Landmarking
replaces one dynamic prediction problem with a sequence of
landmark-specific baseline problems; person-period approaches
introduce a prespecified time grid; and feature-based methods first
compress each longitudinal history into time-fixed summaries.  Our
approach avoids this by formulating the problem around the hazard.
Rather than retrofitting an existing method to time-varying inputs, we
start from a hazard likelihood for predictable covariate processes and
maximize it directly.

This likelihood formulation places the method in the familiar
framework of empirical risk minimization.  The loss functional is the
negative log likelihood, and the hazard estimate minimizes its
empirical average over a function class.  For survival data with predictable
time-dependent covariates, \citet{lee2021boosted} used a fully
nonparametric hazard likelihood and its functional gradient to
construct boosted hazard estimators.  We use the same likelihood
framework but recast the problem within a random forests
architecture.

The resulting methodology has four distinguishing components.  First,
the target is the hazard itself, rather than a relative-risk score, a
discrete-time event probability, or a survival probability on a
prespecified grid.  Second, tree induction uses a node-constant
working hazard under which the empirical risk decomposes into
node-level exposure and event summaries.  This decomposition yields a
split statistic that selects splits by reduction in the likelihood
risk rather than by a heuristic survival impurity.  Third, the
predictable intensity formulation accommodates time-dependent
covariates, including internal covariates.  The covariate state
entering the hazard at time $t$ must be determined by information
available immediately before $t$, enforcing a no-lookahead rule.  This
allows longitudinal laboratory values, vital signs, and treatments to
predict instantaneous event risk without requiring an
external-covariate conditional-survival target.
Fourth, the forest yields pathwise hazard estimates. Once the tree
structures are fixed, we remove the node-constant working restriction
and estimate terminal-node hazards nonparametrically over time. These
estimates define a hazard map for each tree. Following any supplied
predictable covariate process through a tree assembles the
corresponding node-level hazards into a trajectory, and averaging these
trajectories across trees gives the pathwise hazard estimate.

\subsection{Outline}
The remainder of the paper is organized as follows.
Section~\ref{sec:likelihood} reviews survival analysis with
time-varying covariates and develops the hazard likelihood.
Section~\ref{sec:rhf} derives the splitting rule and describes
terminal node estimation.  Section~4 presents algorithmic and
implementation details.  Section~5 evaluates the method on three
time-varying simulations. Section~6 applies the method to a large ICU
case study drawn from the MIMIC-IV
database~\citep{Johnson2023mimiciv,Johnson2023mimicivicu}.  Section~7
concludes with a discussion and directions for future work.

\section{Notation and likelihood}
\label{sec:likelihood}

Survival analysis can be formulated in terms of either the survivor
function or the hazard function.  The survivor function gives the
probability of remaining event-free beyond time $t$, whereas the
hazard gives the instantaneous event rate at time $t$ among those
still at risk.  When covariates are fixed at study entry, these two
descriptions are equivalent and either can serve as the basic target
of inference.  Once covariates are allowed to change over time, the
hazard is the more direct statistical target, because it is local in
time and can be defined from the information available just before
risk is evaluated.  We adopt this local hazard as the primary target.
This section makes that target precise and develops the likelihood
formulation used for tree construction.

\vspace*{-5pt}
\subsection{Time-static covariates}

We begin with the familiar setting in which covariates are fixed over
follow-up.  Each subject is represented by a $p$-dimensional
vector $\X=(X^{(1)},\ldots,X^{(p)})^T$.  Let $T^\ast$ denote the event
time.  For a covariate value $\x$, the conditional survivor function
is $S(t\mid\x)=\PP\{T^\ast>t\mid \X=\x\}$, and the corresponding
continuous-time hazard is
$$
\haz(t,\x)
=
\lim_{\Delta\downarrow 0}\Bigg[
\frac{1}{\Delta}
\PP\{T^\ast\in[t,t+\Delta)\mid T^\ast\ge t,\ \X=\x\}\Bigg].
$$
Thus, $\haz(t,\x)$ describes the local rate of occurrence of the event
at time $t$ among subjects with baseline covariates $\x$ who have
remained event-free up to $t$.

The connection between hazard and survival is seen most directly through
the cumulative hazard,
$$
\Haz(t,\x)=\int_0^t \haz(s,\x)\,ds.
$$
Over a very short interval $[s,s+ds)$, a subject who is still
event-free at time $s$ fails with probability approximately
$\haz(s,\x)\,ds$.  Accumulating these infinitesimal risks from time
$0$ up to time $t$ gives the cumulative hazard, while surviving to time
$t$ means avoiding an event on every such interval.  This leads to the
product-integral identity~\citep{andersen1993},
$$
S(t\mid\x)=\prod_{0\le s<t}\{1-d\Haz(s,\x)\}
$$
which in the continuous case reduces to
$S(t\mid\x)=\exp\{-\Haz(t,\x)\}$.  In the time-static setting,
therefore, the survivor function and the hazard contain the same
information, expressed in two equivalent ways.

\subsection{Time-dependent covariates}
\label{sec:tdc.target}
  
The situation changes once covariates vary with time. Each subject
now contributes a trajectory
$\X(\cdot)=(X^{(1)}(\cdot),\ldots,X^{(p)}(\cdot))^T$. Throughout
Sections~2 and~3, $\X(t)$ denotes the covariate vector entering the
hazard at time $t$, while $\X(\cdot)$ denotes the whole process. We
assume this process is predictable with respect to the observed
history.  This means that the value used at time $t$ is determined only
by information available before $t$. A measurement first
recorded at $t$ therefore cannot affect the hazard at $t$, although it
may affect later hazards. Section~\ref{sec:rhf-startstop} constructs
the predictable routing process that implements this convention.

The statistical target in this setting is the hazard map
$$
(t,\x)\longmapsto\haz(t,\x),
$$
which gives the local event rate at time $t$ for an at-risk subject at
covariate state $\x$. Let $\mathcal F_{t-}$ denote the observed
history just prior to $t$, let $Y(t)\in\{0,1\}$ indicate whether the
subject is at risk just before $t$, and let $N(t)$ count observed
events. We model the dependence of the local event rate on the
history through the current predictable state $\X(t)$. On the at-risk
set,
$$
\PP\{T^\ast\in[t,t+dt)\mid \mathcal F_{t-},\,Y(t)=1\}
=
\haz(t,\X(t))\,dt + o(dt),
$$
or equivalently,
$$
\PP\{dN(t)=1\mid \mathcal F_{t-}\}
=
Y(t)\,\haz(t,\X(t))\,dt + o(dt).
$$
The product $Y(t)\haz(t,\X(t))$ is the predictable intensity of the
event process. Predictability ensures that this intensity uses no
future internal covariate information.

Evaluating the hazard map along $\X(\cdot)$ produces a hazard
trajectory with accumulated hazard
$\int_0^t\haz(s,\X(s))\,ds$. The corresponding exponential,
$$
\exp\!\Bigg\{-\int_0^t \haz(s,\X(s))\,ds\Bigg\},
$$
is a pathwise survivor functional. When the covariate trajectory is
external and specified independently of the event process, this
functional retains the usual conditional-survival interpretation. For
internal covariates, such as longitudinal biomarkers or treatments
tied to a subject's evolving health state, that interpretation does
not generally apply. The realized history through $t$ is observed
only for subjects who remain under observation to $t$, which is the
familiar difficulty of internal time-dependent covariates
\citep[Section~6.3.2]{kalbfleisch2002statistical}. The method
therefore targets the hazard map rather than a pathwise survivor
functional.

The same local intensity argument yields the likelihood in the usual
single-event setting. We reserve $T^\ast$ for the event time and
write $T$ for the observed end of follow-up, whether or not it is an
event time, with $\delta\in\{0,1\}$ indicating whether an event is
observed at $T$. Consider a fine partition
$0=t_0<t_1<\cdots<t_K=T$ and, for $k=0,\ldots,K-1$, set
$\Delta t_k=t_{k+1}-t_k$ and
$\Delta N(t_k)=N(t_{k+1})-N(t_k)$. On each interval,
$\Delta N(t_k)$ is conditionally Bernoulli with
$$
\PP\{\Delta N(t_k)=1\mid\mathcal F_{t_k-}\}
=
Y(t_k)\,\haz(t_k,\X(t_k))\,\Delta t_k+o(\Delta t_k).
$$
Multiplying these contributions and letting the increments tend to
zero gives the likelihood contribution for a single
subject~\citep{gill1990survey}
\Eq
  \exp\!\Bigg\{-\int_0^T Y(s)\,\haz(s,\X(s))\,ds\Bigg\}\,
  \haz(T,\X(T))^{\delta}.
\label{E:hazard-likelihood}
\EndEq

\subsection{Time normalization}
\label{sec:time.normalization}

We train the forest on a time scale normalized to $[0,1]$. 
Let $s$ denote follow-up time on the original scale and set $t=g(s)$,
where $g$ is a fixed increasing transformation into $[0,1]$. For
bounded follow-up, one may take $g(s)=s/\tau$; otherwise, a fixed
bounded monotone transformation may be used. If $\haz_s$ and $\haz_t$
denote the hazards on the original and normalized scales, respectively,
then
$$
\haz_t(g(s),\x)
=
\frac{\haz_s(s,\x)}{g'(s)}.
$$
Hazards estimated on the normalized scale can therefore be returned to
the original scale by the usual change of variables.
This straightforward
conversion does not, however, make the choice of $g$ inconsequential.
Tree induction accumulates
at-risk exposure and evaluates candidate splits on the normalized
scale, so a nonlinear transformation reweights different portions of
follow-up and may alter the node exposure summaries and selected
splits. Thus, $g$ should be considered part of the scale on which the model is fit.
Hereafter, $t$ denotes normalized time, and all likelihood integrals,
split statistics, and terminal-node hazards are defined on $[0,1]$.

\subsection{Likelihood construction}

On this normalized time scale, the observed data for $n$ independent
subjects are $\{(\X_i(\cdot),Y_i(\cdot),N_i(\cdot))\}_{i=1}^n$, and we
write $T_i\in[0,1]$ for the observed end of follow-up for subject $i$.
Here $Y_i(t)\in\{0,1\}$ is the at-risk indicator, equal to one when
subject $i$ is under observation and event-free just prior to time
$t$, and $N_i(t)$ is the counting process of observed events, with
increment $dN_i(t)\in\{0,1\}$ at jump times.  In the familiar
right-censored single-event setting, this becomes
$N_i(t)=\delta_i\,1_{\{T_i\le t\}}$ and $Y_i(t)=1_{\{T_i\ge t\}}$,
where $\delta_i\in\{0,1\}$ indicates whether an event is observed at
$T_i$.  The likelihood contribution in~\eqref{E:hazard-likelihood}
then applies to each subject upon substituting $Y_i$, $N_i$, and $\X_i$.

We parameterize the hazard on the log scale, writing
$F(t,\x)=\log\{\haz(t,\x)\}$ for the log hazard, so that
$\haz(t,\x)=e^{F(t,\x)}$ is positive for any real-valued $F$.  Taking
the logarithm of~\eqref{E:hazard-likelihood} gives the per-subject
log-likelihood
$$
\ell_i(F)
=
\int_0^1 F\bigl(t,\X_i(t)\bigr)\,dN_i(t)
\;-\;
\int_0^1 Y_i(t)\,e^{F(t,\X_i(t))}\,dt,
$$
where the event term $\int_0^1 F(t,\X_i(t))\,dN_i(t)$ reduces to
$\delta_i\,F(T_i,\X_i(T_i))$ in the single-event setting.  When
building a tree, we work with the average negative log-likelihood
over the $n$ subjects,
\Eq
R_n(F)
=
\frac{1}{n}\sum_{i=1}^n\int_0^1 Y_i(t)\,e^{F(t,\X_i(t))}\,dt
\;-\;
\frac{1}{n}\sum_{i=1}^n\int_0^1 F\bigl(t,\X_i(t)\bigr)\,dN_i(t).
\label{E:likelihood.timevary}
\EndEq
By construction, $R_n(F)=-\ell(F)/n$, where
$\ell(F)=\sum_{i=1}^n \ell_i(F)$ is the conditional log likelihood.
Thus $R_n(F)$ can be viewed as a scaled negative conditional log
likelihood.  It can also be viewed as an empirical risk to be minimized
over a function class for $F$, which is the perspective we adopt when
deriving the split rule in the next section.

\section{Tree construction}
\label{sec:rhf}

In a conventional RSF tree, each subject follows a single path from
the root to a terminal node, determined by covariates measured at
baseline. In RHF, this path evolves with the covariate process. At
each time $t$, the current predictable covariate state $\X_i(t)$ is
routed through the tree, so the same subject may contribute at-risk
exposure to different nodes as the path changes. If an event occurs,
it is assigned to the node occupied at that time. This dynamic routing
mirrors the likelihood in Section~\ref{sec:likelihood}, which separates
at-risk exposure from event increments along the path. Aggregating
these quantities within candidate tree regions produces the node-level
summaries used to evaluate splits by empirical-risk reduction.

We separate the hazard model used to grow a tree from the
model used for the hazard map. During tree induction, a best-first search
constructs the partition from admissible covariate splits; time itself
is not used as a splitting variable. Candidate splits are evaluated
under a working model that assigns a time-constant log hazard to each
current node. Under this restriction, a node's contribution to the
empirical risk depends only on its total at-risk exposure and event
total. Theorem~\ref{T:empR} gives the corresponding nodewise risk
decomposition and the resulting closed-form split criterion. Starting
from the root, the algorithm repeatedly selects the admissible split with the
largest risk reduction among all current leaves and stops when no
admissible split remains. Once the partition is fixed, the working
restriction is removed, and a time-varying hazard is estimated in
each terminal node by minimizing empirical risk on
a grid of evaluation times. 

This construction yields a single tree. For any predictable
covariate process, each tree routes the current state at time $t$ to a
terminal node and returns that node's estimated hazard at
$t$. Averaging these values across trees yields the pathwise
hazard estimate along the supplied covariate path. The next two
subsections develop the two tree-level ingredients of this estimator:
the likelihood-based split criterion and the terminal-node hazard
estimator. Section~4 then assembles them into a complete algorithm for
training and test time estimation and describes the associated
implementation details.

\subsection{Empirical risk minimization for tree splits}

Given a current tree, let $\Pi=\{A_1,A_2,\ldots\}$ denote its leaves,
which form a partition of $\RR^p$.  We evaluate candidate splits
over log hazards of the form
\Eq
F(t,\x) = \sum_{A_j\in\Pismall} c_j\,1_{\{\x\in A_j\}},
\label{E:F-tree-partition}
\EndEq
which assigns a single working log hazard $c_j$ to each current leaf.
For this class, the at-risk, event, and covariate processes
in~\eqref{E:likelihood.timevary} enter the empirical risk through two
node-level summaries.

For a given node $A_j$, define the two summary quantities
$$
U_j
=
\frac{1}{n}
\sum_{i=1}^n
\int_0^1 Y_i(t)\,1_{\{\X_i(t)\in A_j\}}\,dt,
\qquad
V_j
=
\frac{1}{n}
\sum_{i=1}^n
\int_0^1 1_{\{\X_i(t)\in A_j\}}\,dN_i(t).
$$
Here $U_j$ is the average amount of time the sample spends at risk with
covariates falling in $A_j$, and $V_j$ is the average number of events
that occur while $\X_i(t)\in A_j$.  In the single-event setting, $V_j$
is the proportion of observed events whose covariates at the event time
lie in $A_j$.  These two quantities, total exposure and total events
within a node, are the only summaries needed to evaluate a candidate
split.

The next result is the tree version of the nonparametric hazard
likelihood risk used in \citet{lee2021boosted}.

\begin{theorem}\label{T:empR}
Let $F$ be of the form~\eqref{E:F-tree-partition}. Then the empirical
risk~\eqref{E:likelihood.timevary} can be written as
\Eq
R_n(F) = \sum_{j:\,U_j>0}\Bigl(e^{c_j}\,U_j-c_j\,V_j\Bigr).
\label{E:treeR}
\EndEq
Within a fixed partition $\{A_j\}$, the contribution of each region $A_j$
depends only on $c_j$, $U_j$, and $V_j$.
\end{theorem}

\subsubsection{Initialization step}

Theorem~\ref{T:empR} is the basis for split selection.  At the root
there is a single node $A_1=\RR^p$, so the working log hazard is
constant over the entire covariate space, $F_0(t,\x)=c_0$.  In the
notation of Theorem~\ref{T:empR}, the risk depends on one pair
$(U_0,V_0)$, where
$$
U_0
=
\frac{1}{n}\sum_{i=1}^n \int_0^1 Y_i(t)\,dt,
\qquad
V_0
=
\frac{1}{n}\sum_{i=1}^n \int_0^1 dN_i(t).
$$
The quantity $U_0$ is the average time subjects are at risk and $V_0$
is the average number of observed events, which reduces to the event
proportion in the single-event setting.  For $F_0$, \eqref{E:treeR}
becomes $R_n(F_0)=e^{c_0}U_0-c_0V_0$.  When $V_0>0$, the minimizer is
$c_0=\log(V_0/U_0)$.  The root working hazard is therefore the overall
empirical hazard $V_0/U_0$, total events divided by total exposure
time.

\subsubsection{Induction step}

After $s$ splits, the tree induces a partition
$\{A_1,\ldots,A_{s+1}\}$ of $\RR^p$ and a working log hazard
$F_s(t,\x)=\sum_{j=1}^{s+1} c_j\,1_{\{\x\in A_j\}}$.
To obtain the next split, consider splitting a region $A_j$ into two
children $A_{j,1}$ and $A_{j,2}$ and updating
$$
F_{s+1}(t,\x)
=
F_s(t,\x)
-
c_j\,1_{\{\x\in A_j\}}
+
c_{j,1}\,1_{\{\x\in A_{j,1}\}}
+
c_{j,2}\,1_{\{\x\in A_{j,2}\}}.
$$
For a candidate child region $A_{j,\ell}$, $\ell=1,2$, define
$$
U_{j,\ell}
  = \frac{1}{n}
     \sum_{i=1}^n
     \int_0^1 Y_i(t)\,1_{\{\X_i(t)\in A_{j,\ell}\}}\,dt,
\qquad
V_{j,\ell}
  = \frac{1}{n}
     \sum_{i=1}^n
     \int_0^1 1_{\{\X_i(t)\in A_{j,\ell}\}}\,dN_i(t).
$$
The ratio $V_{j,\ell}/U_{j,\ell}$ is the empirical hazard in
$A_{j,\ell}$, computed as events divided by exposure time within that
child.

By Theorem~\ref{T:empR}, the contribution of $A_{j,\ell}$ to $R_n(F)$
is $e^{c_{j,\ell}}U_{j,\ell}-c_{j,\ell}V_{j,\ell}$.  When
$U_{j,\ell}>0$ and $V_{j,\ell}>0$, this expression is minimized at
$c_{j,\ell} = \log(V_{j,\ell}/U_{j,\ell})$.  The parent region $A_j$
satisfies $U_j = U_{j,1}+U_{j,2}$, $V_j = V_{j,1}+V_{j,2}$, and its
working log hazard is
$$
c_j
=
\log\!\left(\frac{V_j}{U_j}\right)
=
\log\!\left(\frac{V_{j,1}+V_{j,2}}{U_{j,1}+U_{j,2}}\right),
$$
which is the empirical hazard obtained by pooling the two children.

\subsubsection{Split statistic}

Substitute the optimal values $c_j$ and $c_{j,\ell}$
into~\eqref{E:treeR} and use $U_j=U_{j,1}+U_{j,2}$ and
$V_j=V_{j,1}+V_{j,2}$.  The drop in empirical risk from splitting
$A_j$ into $A_{j,1}$ and $A_{j,2}$ is then
\Array
R_n(F_s) - R_n(F_{s+1})
&=&
V_{j,1}\log\!\left(\frac{V_{j,1}}{U_{j,1}}\right)
+
V_{j,2}\log\!\left(\frac{V_{j,2}}{U_{j,2}}\right)
-
V_j\log\!\left(\frac{V_j}{U_j}\right).
\EndArray
For a fixed parent node $A_j$, the last term does not depend on the
choice of cut, so the best cut within $A_j$ maximizes the sum of the
two child terms.  The best-first growth rule then compares the best
achievable risk reduction across the current leaves and applies the
split with the largest reduction.

\begin{definition}\label{D:RHF.splitrule}\emph{
The split statistic for splitting a node $A_j$ into regions
$A_{j,1}$ and $A_{j,2}$ is
\Eq
S(A_{j,1},A_{j,2})
=
V_{j,1}\log\!\left(\frac{V_{j,1}}{U_{j,1}}\right)
+
V_{j,2}\log\!\left(\frac{V_{j,2}}{U_{j,2}}\right),
\label{E:RHF.split.stat}
\EndEq
and the associated empirical risk reduction is
\Eq
\D_n(A_{j,1},A_{j,2})
=
V_{j,1}\log\!\left(\frac{V_{j,1}}{U_{j,1}}\right)
+
V_{j,2}\log\!\left(\frac{V_{j,2}}{U_{j,2}}\right)
-
(V_{j,1}+V_{j,2})\log\!\left(\frac{V_{j,1}+V_{j,2}}{U_{j,1}+U_{j,2}}\right).
\label{E:RHF.risk.reduction}
\EndEq
If $R_n(F_s)$ is the empirical risk for the current working log hazard
and $R_n(F_{s+1})$ is the risk after the split, then
$R_n(F_{s+1})=R_n(F_s)-\D_n(A_{j,1},A_{j,2})$.}
\end{definition}

Each ratio $V_{j,\ell}/U_{j,\ell}$ is the empirical hazard in child
$A_{j,\ell}$, while $(V_{j,1}+V_{j,2})/(U_{j,1}+U_{j,2})$ is the
pooled hazard in the parent.  The quantity
$\D_n(A_{j,1},A_{j,2})$ therefore measures the improvement in fit
from allowing the children different hazard levels rather than
forcing them to share the parent value.  This plays the same role as
impurity reduction in classification trees, but here the objective is
empirical risk reduction of the hazard
likelihood~\eqref{E:likelihood.timevary}.

\subsection{Terminal node estimator}

To obtain the terminal node hazard estimator,
let $\Pi$ denote the partition generated from the tree induction, and fix a
time grid $\tt=\{t_0=0<t_1<\cdots<t_K=1\}$, with intervals
$I_k=(t_{k-1},t_k]$.  The interior grid points for the evaluation grid $\tt$
are selected from
the observed event times.  The terminal node estimator is chosen to
minimize the
empirical risk over the class of log hazards
$$
F_{\tt}(t,\x)
=
\sum_{A_j\in\Pismall}\sum_{k=1}^K
c_{j,k}\,1_{\{\x\in A_j\}}\,1_{\{t\in I_k\}}.
$$
This refines each terminal node $A_j$ into the node-time cells
$A_j\times I_k$.  Applied to these cells, Theorem~\ref{T:empR} gives
$$
R_n(F_{\tt})
=
\sum_{A_j\in\Pismall}\sum_{k=1}^K
\left(e^{c_{j,k}}U_{j,k}-c_{j,k}V_{j,k}\right),
$$
where
$$
U_{j,k}
=
\frac{1}{n}\sum_{i=1}^n
\int_{t_{k-1}}^{t_k}
Y_i(t)\,1_{\{\X_i(t)\in A_j\}}\,dt,
\qquad
V_{j,k}
=
\frac{1}{n}\sum_{i=1}^n
\int_{t_{k-1}}^{t_k}
1_{\{\X_i(t)\in A_j\}}\,dN_i(t).
$$
Here $U_{j,k}$ is the average at-risk exposure contributed to node
$A_j$ during $I_k$, and $V_{j,k}$ is the corresponding average number
of events.

For $U_{j,k}>0$, the cell contribution is minimized at
$c_{j,k}=\log(V_{j,k}/U_{j,k})$ when $V_{j,k}>0$, and the minimizing
hazard is zero when $V_{j,k}=0$.  The terminal-node hazard estimator
is therefore
\Eq
\hhat_{A_j}(t)
=
\frac{V_{j,k}}{U_{j,k}}\,1_{\{U_{j,k}>0\}},
\qquad t\in(t_{k-1},t_k],\qquad k=1,\ldots,K.
\label{E:tnh}
\EndEq
The estimate is the number of events observed in the node--time cell
divided by its total at-risk exposure.  It is the time-resolved
counterpart of the node value $V_j/U_j$ used during tree induction, with
$U_j=\sum_{k=1}^K U_{j,k}$ and $V_j=\sum_{k=1}^K V_{j,k}$.

For each tree, the collection $\{\hhat_{A_j}(t)\}$ defines a hazard
function over node region and time.  Ensembling the resulting
tree-level functions yields the estimated hazard map, the primary
statistical target of the method.  Evaluating this map along a predictable
covariate process produces a case-specific hazard trajectory.
The next section develops this
construction.

\begin{remark}
The evaluation grid $\{t_0=0<t_1<\cdots<t_K=1\}$ affects only the temporal resolution of the
terminal-node hazard estimator.  It is introduced \emph{after} the tree
partition has been determined, so the choice of $K$ does not play a
role in tree growth.  The interior grid points are selected from the
observed event times.  Retaining all event times gives the finest
available event-time resolution, while retaining a subset pools
neighboring intervals along with their exposure and event summaries.
Thus $K$ trades temporal detail for stability and computation, and may
be selected adaptively, for example using out-of-bag risk
(Section~\ref{sec:rhf-oob}).

The interpretation becomes especially clear as the grid is refined.
The intensity of $N(t)$ is $Y(t)\haz(t,\X(t))$, so
$V_{j,k}/U_{j,k}$ estimates the exposure-weighted average of
$\haz(t,\X(t))$ over the node-time region $A_j\times I_k$.  On a
coarser grid this average is taken over a longer time interval.  As the
intervals become shorter, the average localizes in time and approaches
the risk-set weighted node hazard
$$
\haz_{A_j}(t)
=
\frac{
\E\left[
Y(t)\,1_{\{\X(t)\in A_j\}}\,
\haz(t,\X(t))
\right]
}{
\E\left[
Y(t)\,1_{\{\X(t)\in A_j\}}
\right]
}.
$$
This is the average instantaneous event rate among subjects who are
still at risk at time $t$ and whose predictable covariates place them
in $A_j$.
\end{remark}

\begin{remark}
A further consequence of~\eqref{E:tnh} is a generalized conservation
of events property.  Because the terminal nodes partition the
covariate-time space  and $\hhat_{A_j}(t)$ is constant on each
$I_k$,
$$
\begin{aligned}
\frac{1}{n}\sum_{i=1}^n
\int_0^1
Y_i(t)
\left\{
\sum_{A_j\in\Pismall}
\hhat_{A_j}(t)1_{\{\X_i(t)\in A_j\}}
\right\}dt
&=
\sum_{A_j\in\Pismall}\sum_{k=1}^K
U_{j,k}\hhat_{A_j}(t_k)\\
&=
\sum_{A_j\in\Pismall}\sum_{k=1}^K V_{j,k}
=
\frac{1}{n}\sum_{i=1}^n\int_0^1 dN_i(t)
=
V_0.
\end{aligned}
$$
Thus the total exposure-weighted hazard, multiplied by $n$,
equals the total number of observed events.
In the time-static, right-censored setting, the property reduces to a
known identity.  Here $\X_i(t):=\X_i$ and subject $i$ remains in a
single terminal node, denoted by $A(i)$.  Let
$\Hhat_{A_j}(t) = \int_0^t\hhat_{A_j}(s)\,ds$,
then the generalized identity becomes
$$
\sum_{i=1}^n
\Hhat_{A(i)}(T_i)
=
\sum_{i=1}^n
\int_0^1Y_i(t)\hhat_{A(i)}(t)\,dt
=
\sum_{i=1}^n
\int_0^1dN_i(t)
=
\sum_{i=1}^n\d_i.
$$
Thus the sum of the 
cumulative hazards evaluated at the observed follow-up times equals the
total number of events.  This is the conservation of events
principle introduced for RSF in \citet{ishwaran2008rsf}.
\end{remark}

\section{Algorithm and implementation details}
\label{sec:oper.rhf}

The previous section developed the likelihood-based split criterion
and terminal-node hazard estimator, the two ingredients needed to
construct a single tree.  We now operationalize these theoretical
pieces by introducing a start--stop representation of each subject's
predictable covariate trajectory.  Under this representation, the
exposure and event contributions appearing in the split criterion
reduce exactly to finite sums over observed data rows, allowing the
theoretical splitting rule to be evaluated directly and efficiently
during tree growth.  Once the forest is trained, a pathwise assembly
rule uses the same intervalwise structure to evaluate the forest along
any supplied predictable covariate process and produce a case-specific
hazard estimate.

Algorithm~\ref{alg:rhf} summarizes the resulting procedure.  Using
the start--stop preprocessing described in
Section~\ref{sec:rhf-startstop}, the algorithm grows each tree from a
subject-level subsample (line~2) using a best-first search over
admissible covariate splits (lines~5--8).  At each iteration, it
evaluates candidate splits across the current leaves using the
finite-sum calculations in Section~\ref{sec:rhf-split-eval} and applies
the split with the largest empirical risk reduction.  Candidate splits
act on covariates, not on time, and tree growth ends after $\ss$ splits
or when no admissible split remains.  Once the partition is fixed, the
terminal-node step (line~10) removes the node-constant working
restriction used during induction and estimates a time-varying hazard
in each terminal node on the evaluation grid $\tt$ with
estimator~\eqref{E:tnh}.  At test time (line~13), the forest routes a
supplied predictable covariate process through each tree and averages
the resulting hazard trajectories, as developed in
Section~\ref{sec:rhf-prediction}.
Sections~\ref{sec:rhf-oob} and~\ref{sec:rhf-vimp} describe
out-of-bag estimation and a post-hoc variable priority procedure,
respectively.

\begin{figure}
\vspace*{2in}

\begin{algorithm}[H]
\caption{Random Hazard Forests (RHF)}
\label{alg:rhf}
\begin{algorithmic}[1]

\vspace*{5pt}
\item[]\hspace*{-\leftmargin}\textbf{Input}
\vspace*{2pt}

Training data
$\{(\X_i(\cdot),Y_i(\cdot),N_i(\cdot))\}_{i=1}^n$;
number of trees $B$; number of candidate covariates considered at
each leaf \texttt{mtry}; maximum number of splits per tree $\ss$;
evaluation grid $\tt$.

\vspace*{5pt}
\item[]\hspace*{-\leftmargin}\textbf{Preprocessing}
\vspace*{2pt}

Rescale all observed times and the evaluation grid $\tt$ to a common
$[0,1]$ scale (Section~\ref{sec:time.normalization}). Encode each
subject's observed history in start--stop form as described in
Section~\ref{sec:rhf-startstop}, yielding intervals
$(S_{i,r},T_{i,r}]$ with event increments $\Delta N_{i,r}$ and
predictable routing states $\Z_{i,r}$, $r=1,\ldots,n_i$,
$i=1,\ldots,n$.

\vspace*{10pt}
\item[]\hspace*{-\leftmargin}\textbf{Training}
\vspace*{3pt}

\FOR{$b = 1$ to $B$}
  \STATE Draw a subject-level subsample of size
         $\lfloor 0.632\,n\rfloor$ without replacement, retaining the
         full trajectory of each selected subject. Let $\oo_b$ denote
         the subjects not selected and store it for out-of-bag
         estimation.
                  
  \STATE Initialize the root node with the selected subjects and set the
         current partition to $\Pi_b=\{\RR^p\}$.

  \FOR{$s = 1$ to $\ss$}
    \STATE For each current leaf $A\in\Pi_b$, sample \texttt{mtry}
      covariates without replacement. For each admissible cutpoint on a
      sampled covariate, form child nodes $A_1$ and $A_2$ and compute
      the child aggregates $(U_{A,\ell},V_{A,\ell})$, $\ell=1,2$.
    \STATE Evaluate the empirical-risk reduction
      $\D_n(A_1,A_2)$ in~\eqref{E:RHF.risk.reduction} for each
      candidate split across all current leaves.
    \STATE If no admissible candidate remains, stop tree growth.
      Otherwise, select the split with the largest empirical-risk
      reduction.
    \STATE Apply the selected split by replacing its parent leaf with
      $A_1$ and $A_2$, and update $\Pi_b$.
  \ENDFOR

  \STATE Estimate the time-varying hazard
       $\hhat_{A,b}(t)$ for each terminal node $A\in\Pi_b$ on the grid
       $\tt$ using~\eqref{E:tnh}.
\ENDFOR

\STATE The trained forest is the collection
$\{(\Pi_b,\{\hhat_{A,b}(t):A\in\Pi_b\})\}_{b=1}^B$, 
which define the forest hazard map
$$
\hhat_{\rm RHF}(t,\x)
=
\frac{1}{B}
\sum_{b=1}^B
\sum_{A\in\Pi_b}
\hhat_{A,b}(t)1_{\{\x\in A\}}.
$$

\vspace*{10pt}
\item[]\hspace*{-\leftmargin}\textbf{Pathwise evaluation}
\vspace*{3pt}

\STATE For a supplied predictable covariate process $\X(\cdot)$ on the
same time scale, evaluate the forest map along the path.
Operationally, route the current covariate state through each tree and
average the corresponding terminal-node hazards to obtain
the pathwise hazard estimate,
$\hhat_{\rm RHF}(t,\X(t))$.

\vspace*{5pt}

\end{algorithmic}
\end{algorithm}

\vspace*{2in}
\end{figure}

\subsection{Start--stop representation of evolving covariate paths}
\label{sec:rhf-startstop}

For subject $i$, we represent the predictable covariate process
$\X_i(\cdot)$ and the at-risk and event processes $Y_i(\cdot)$ and
$N_i(\cdot)$ in start--stop form.  Each row corresponds to an interval
over which the routing covariate state is fixed and records the event
increment over that interval.  Let $\mathcal D_i$ denote the set of
follow-up times covered by these intervals,
\Eq
\mathcal D_i
=
\bigcup_{r=1}^{n_i}(S_{i,r},T_{i,r}]
\subseteq (0,1].
\label{E:time.set.i}
\EndEq
The intervals in~\eqref{E:time.set.i} are disjoint and ordered as
$$
0\le S_{i,1}<T_{i,1}\le S_{i,2}<T_{i,2}\le\cdots\le
S_{i,n_i}<T_{i,n_i}\le 1.
$$
Here $S_{i,r}$ and $T_{i,r}$ are the start and stop times of the $r$th
row. A new row begins whenever the routing covariate state changes.  The
encoding may split rows further at event times or at-risk transitions, so that
each interval receives the appropriate exposure and event
contributions.  A stop time $T_{i,r}$ therefore marks a row boundary,
not necessarily an event.
The ordering permits gaps in the supplied path when
$S_{i,r+1}>T_{i,r}$.  If follow-up is observed continuously from the
origin through a final endpoint $T_i$, then $\mathcal D_i=(0,T_i]$,
  with $S_{i,1}=0$, $S_{i,r+1}=T_{i,r}$ and $T_{i,n_i}=T_i$.

The resulting start--stop record has one row for each interval:
$$
{\setlength{\arraycolsep}{10pt}\renewcommand{\arraystretch}{1.3}
\begin{array}{cccc}
  \text{subject} & (\text{start},\text{stop}] & \Delta N
  & \text{routing covariate} \\
  \specialrule{.2em}{.1em}{.1em}\\[-10pt]
  i & (S_{i,1},T_{i,1}] & \Delta N_{i,1} & \Z_{i,1} \\
  \vdots & \vdots & \vdots & \vdots \\
  i & (S_{i,n_i},T_{i,n_i}] & \Delta N_{i,n_i} & \Z_{i,n_i}.
\end{array}}
$$
For row $r$, we define
$$
\Delta N_{i,r}
=
N_i(T_{i,r})-N_i(S_{i,r}),
\qquad
\Z_{i,r}
=
\X_i(T_{i,r}-).
$$
Here $\Delta N_{i,r}$ is the event increment over the interval, and
$\Z_{i,r}$ is the predictable covariate state used for tree routing.
The left limit defining $\Z_{i,r}$ ensures that an event at $T_{i,r}$
is evaluated using only information available immediately before that
time.  Because $\Z_{i,r}$ is fixed over $(S_{i,r},T_{i,r}]$, it
determines which node receives the interval's at-risk exposure and,
when $\Delta N_{i,r}=1$, its event increment.  Equivalently, the rows
define the piecewise-constant routing process
$$
\Z_i(t)
=
\sum_{r=1}^{n_i}
\Z_{i,r}\,1_{\{t\in(S_{i,r},T_{i,r}]\}},
\qquad t\in\mathcal D_i.
$$

For test time evaluation, the start--stop formulation is slightly
different.  A path supplied at test time consists of predictable
covariate states alone, with no event or at-risk process.  Its
intervals are therefore determined only by changes in the routing
state.  If $\X(\cdot)$ is the supplied predictable covariate process
with fixed states $\X(t)=\x_r$ on $(S_r,T_r]$ for $r=1,\ldots,m$, then
the routing process is $\Z(t) = \sum_{r=1}^{m} \x_r \,1_{\{t\in(S_r,T_r]\}}$
with time domain $t\in \mathcal D_X=\bigcup_{r=1}^m(S_r,T_r]$.

\subsection{Evaluating a candidate split}
\label{sec:rhf-split-eval}

Each candidate split is scored by the empirical risk
reduction~\eqref{E:RHF.risk.reduction}, computed from the split
statistic~\eqref{E:RHF.split.stat}.  The only inputs needed for each child
are its exposure and event aggregates $(U,V)$, which under the start--stop
representation reduce to finite sums over intervals.

Consider splitting node $A_j$ on coordinate $X^{(k)}$ at cutpoint $c$.  The
children are
$A_{j,1}=A_j\cap\{X^{(k)}\le c\}$ and
$A_{j,2}=A_j\cap\{X^{(k)}>c\}$.  Their interval memberships are
$$
I_{i,r}(A_{j,1})
=
I_{i,r}(A_j)1_{\{Z_{i,r}^{(k)}\le c\}},
\qquad
I_{i,r}(A_{j,2})
=
I_{i,r}(A_j)1_{\{Z_{i,r}^{(k)}>c\}},
$$
where membership of interval $(S_{i,r},T_{i,r}]$ for node $A_j$ is denoted by
$I_{i,r}(A_j)=1_{\{\Z_{i,r}\in A_j\}}$.

For $\ell=1,2$, the corresponding exposure and event aggregates are
$$
U_{A_{j,\ell}}
=
\frac{1}{n}\sum_{i=1}^n\sum_{r=1}^{n_i}
\left\{\int_{S_{i,r}}^{T_{i,r}}Y_i(t)\,dt\right\}
I_{i,r}(A_{j,\ell}),
\qquad
V_{A_{j,\ell}}
=
\frac{1}{n}\sum_{i=1}^n\sum_{r=1}^{n_i}
\Delta N_{i,r} \, I_{i,r}(A_{j,\ell}),
$$
which can be calculated without the need for numerical integration.
Because $\Z_i(t)$ is constant on each interval $(S_{i,r},T_{i,r}]$, the
indicator $1_{\{\Z_i(t)\in A\}}$ is also constant there and equals
$I_{i,r}(A)$.  Consequently, the finite sums above agree with the integrals
in Section~\ref{sec:rhf}:
$$
\begin{aligned}
\int_0^1Y_i(t)1_{\{\Z_i(t)\in A\}}\,dt
&=
\sum_{r=1}^{n_i}
\left\{\int_{S_{i,r}}^{T_{i,r}}Y_i(t)\,dt\right\}I_{i,r}(A),\\[6pt]
\int_0^1 1_{\{\Z_i(t)\in A\}}\,dN_i(t)
&=
\sum_{r=1}^{n_i}\Delta N_{i,r} \, I_{i,r}(A).
\end{aligned}
$$
Once $(U_{A_{j,\ell}},V_{A_{j,\ell}})$ have been computed, they enter
directly into the split statistic~\eqref{E:RHF.split.stat} and the risk
reduction~\eqref{E:RHF.risk.reduction}, which quantify the improvement in
fit obtained by replacing $A_j$ with $A_{j,1}$ and $A_{j,2}$.

\subsection{Case-specific hazard from a trained forest}
\label{sec:rhf-prediction}

Once tree $b$ is grown, each terminal node $A\in\Pi_b$ contains a
time-varying hazard estimate $\hhat_{A,b}(t)$
from~\eqref{E:tnh}. Because the terminal nodes partition the
covariate space, these estimates define the tree-level hazard map
$$
\hhat_b(t,\x)
=
\sum_{A\in\Pi_b}
\hhat_{A,b}(t)1_{\{\x\in A\}}.
$$
At a fixed pair $(t,\x)$, this map returns the hazard at time $t$ from
the terminal node containing $\x$.  The central object of RHF is the
forest hazard map obtained by averaging these tree-level maps,
$$
\hhat_{\rm RHF}(t,\x)
=
\frac{1}{B}\sum_{b=1}^B \hhat_b(t,\x).
$$
This is the forest estimator of the target hazard map
$(t,\x)\mapsto\haz(t,\x)$ introduced in
Section~\ref{sec:tdc.target}.

To obtain a case-specific hazard
trajectory, let $\X(\cdot)$ be a supplied predictable covariate
process with time domain
$\mathcal D_X=\bigcup_{r=1}^m(S_r,T_r]$
using the start--stop representation of
Section~\ref{sec:rhf-startstop} with
$\X(t)=\x_r$ on $(S_r,T_r]$. Evaluating tree $b$ along this path
gives
$$
\hhat_b(t,\X(t))
=
\sum_{r=1}^{m}
\hhat_b(t,\x_r)1_{\{t\in(S_r,T_r]\}} 
=
\sum_{r=1}^{m}\sum_{A\in\Pi_b}
\hhat_{A,b}(t)1_{\{\x_r\in A\}}\,
1_{\{t\in(S_r,T_r]\}},
\qquad t\in\mathcal D_X.
$$
On each interval, the current covariate state selects a terminal node,
whose estimated hazard is followed until the state changes.

Averaging the tree-specific trajectories pointwise yields the
case-specific forest hazard trajectory,
$$
\hhat_{\rm RHF}(t,\X(t))
=
\frac{1}{B}\sum_{b=1}^B
\hhat_b(t,\X(t)) \\
=
\sum_{r=1}^m
\hhat_{\rm RHF}(t,\x_r)
1_{\{t\in(S_r,T_r]\}},
\qquad t\in\mathcal D_X.
$$
The first expression averages the trajectories assembled within each
tree; the second shows that the result is exactly the forest hazard map
evaluated along the supplied predictable covariate path.

\begin{remark}
The training-data case is a special case of the preceding construction.
For subject $i$, the hazard maps are evaluated along the predictable
routing process $\Z_i(t)$ from Section~\ref{sec:rhf-startstop}, defined
on the time domain $\mathcal D_i$ in~\eqref{E:time.set.i}.
Substituting $(S_{i,r},T_{i,r}]$ and $\Z_{i,r}$ into the preceding
display gives the tree-specific hazard trajectory
\Eq
\hhat_{i,b}(t)
=
\hhat_b(t,\Z_i(t))
=
\sum_{r=1}^{n_i}\sum_{A\in\Pi_b}
\hhat_{A,b}(t)
1_{\{\Z_{i,r}\in A\}}\,
1_{\{t\in(S_{i,r},T_{i,r}]\}},
\qquad t\in\mathcal D_i.
\label{E:case-hazard-tree}
\EndEq
Averaging these trajectories gives the in-sample forest estimate
$$
\hhat_i^{\,\mathrm{RHF}}(t)
=
\frac{1}{B}\sum_{b=1}^B\hhat_{i,b}(t) 
=
\hhat_{\rm RHF}(t,\Z_i(t)) 
=
\sum_{r=1}^{n_i}
\hhat_{\rm RHF}(t,\Z_{i,r})
1_{\{t\in(S_{i,r},T_{i,r}]\}},
\qquad t\in\mathcal D_i.
$$
Thus, the training-data hazard trajectory is the forest hazard
map evaluated along the subject's observed predictable covariate path.
\end{remark}

\begin{remark}
The forest also provides a case-specific cumulative-hazard estimator along a
supplied covariate path, defined by
$$
\Hhat_b(t,\X(\cdot))
=
\int_{(0,t]\cap\mathcal D_X}
\hhat_b(u,\X(u))\,du.
$$
Thus the cumulative hazard is formed by accumulating the node-level
hazard increments encountered along the covariate path.  If
$\mathcal D_X=(0,1]$, this is the usual full-horizon cumulative hazard;
otherwise it is accumulated only over the portions of time on which the
path has been specified.  Averaging over trees gives the forest-level
cumulative-hazard estimator
$$
\Hhat_{\rm RHF}(t,\X(\cdot))
=
\frac{1}{B}\sum_{b=1}^B \Hhat_b(t,\X(\cdot)).
$$
\end{remark}

\subsection{Out-of-bag estimation and empirical risk}
\label{sec:rhf-oob}

For each tree $b=1,\ldots,B$, let $\oo_b$ denote the set of subjects
not included in the training subsample for that tree.  Out-of-bag
(OOB) estimation uses only these trees, so each subject is evaluated
by a forest that was fit without using that subject's data.

\subsubsection*{OOB case-specific hazard}

For tree $b$, let $\hhat_{i,b}(t)$ denote the
case-specific hazard for subject $i$ defined in~\eqref{E:case-hazard-tree}.
The OOB ensemble hazard for subject $i$ averages only over trees for
which $i$ was not in the training subsample,
$$
\hhat_i^{\mathrm{oob}}(t)
=
\frac{
\sum_{b=1}^B \hhat_{i,b}(t)1_{\{i\in\oo_b\}}
}{
\sum_{b=1}^B 1_{\{i\in\oo_b\}}
},
\qquad
t\in\mathcal D_i .
$$
Each term comes from a tree fit without subject $i$, so the OOB
ensemble provides an internally cross-validated case-specific
estimate.

\subsubsection*{OOB empirical risk}

The OOB empirical risk uses the same likelihood loss as the empirical
risk, but evaluates each subject using only trees for which that
subject was out of bag.  Writing $\haz_i(t)=e^{F(t,\X_i(t))}$, the
empirical risk can be written as
\Eq
R_n(\haz)
=
\frac{1}{n}\sum_{i=1}^n\int_0^1Y_i(t)\haz_i(t)\,dt
-
\frac{1}{n}\sum_{i=1}^n\int_0^1\log\{\haz_i(t)\}\,dN_i(t).
\label{E:Rn.h}
\EndEq

The corresponding OOB contribution for subject $i$ is
evaluated by substituting the case-specific OOB hazard
$\hhat_i^{\mathrm{oob}}(t)$ into this loss on the time set where the
subject's covariate path is defined
$$
\widehat R_i^{\mathrm{oob}}
=
\int_{\mathcal D_i}Y_i(t)\hhat_i^{\mathrm{oob}}(t)\,dt
-
\int_{\mathcal D_i}
\log\{\hhat_i^{\mathrm{oob}}(t)\}\,dN_i(t).
$$

To compute this quantity on the evaluation grid, let
$I_k=(t_{k-1},t_k]$ and define the row--grid overlap set
$$
\mathcal G_i
=
\{(r,k):(S_{i,r},T_{i,r}]\cap I_k\ne\emptyset\}.
$$
For $(r,k)\in\mathcal G_i$, define
$$
L_{i,r,k}
=
\int_{t_{k-1}}^{t_k}
1_{\{u\in(S_{i,r},T_{i,r}]\}}Y_i(u)\,du,
\qquad
D_{i,r,k}
=
\int_{t_{k-1}}^{t_k}
1_{\{u\in(S_{i,r},T_{i,r}]\}}\,dN_i(u).
$$
Also write $\hhat_{i,r,k}^{\mathrm{oob}}$ for the OOB grid value used
on this row--grid overlap, namely
$$
\hhat_{i,r,k}^{\mathrm{oob}}
=
\frac{
\sum_{b=1}^B
\sum_{A\in\Pi_b}
\hhat_{A,b}(t_k)1_{\{\Z_{i,r}\in A\}}1_{\{i\in\oo_b\}}
}{
\sum_{b=1}^B1_{\{i\in\oo_b\}}
}.
$$
The finite-sum OOB risk contribution is then
$$
\widehat R_i^{\mathrm{oob}}
=
\sum_{(r,k)\in\mathcal G_i}
\hhat_{i,r,k}^{\mathrm{oob}}L_{i,r,k}
-
\sum_{(r,k)\in\mathcal G_i}
D_{i,r,k}\log\{\hhat_{i,r,k}^{\mathrm{oob}}\}.
$$
Aggregating over subjects gives the ensemble OOB risk,
$$
\widehat R^{\mathrm{oob}}
=
\frac{1}{n}\sum_{i=1}^n\widehat R_i^{\mathrm{oob}}.
$$
This is a cross-validated analogue of the empirical risk
$R_n(\haz)$ in~\eqref{E:Rn.h} and can be used to monitor convergence
or compare forests fit with different tuning parameters.

\subsection{Time-localized variable priority}
\label{sec:rhf-vimp}

The forest also supports a post hoc, time-local variable selection
procedure. This method adapts the rule-release variable priority
framework of~\cite{varpro,ishwaran2025multivariate} from subject-level
baseline records to the start--stop intervals of
Section~\ref{sec:rhf-startstop}. In this
approach, each interval represents a local covariate state, and the
diagnostic assesses whether removing the effect of a variable by
releasing it from a forest rule changes the hazard.

Each terminal node corresponds to a root-to-node decision rule. Let
$\zeta$ denote such a rule and let $r(\zeta)\subset\RR^p$ be its rule
region. For coordinate $\nu\in\{1,\ldots,p\}$, let $r^{(-\nu)}(\zeta)$ be the region
obtained by removing all split constraints involving $X^{(\nu)}$ while
leaving the remaining constraints unchanged. The corresponding
near-miss region from this release operation is
$$
r^{\mathrm{miss}}_\nu(\zeta)
=
r^{(-\nu)}(\zeta)\setminus r(\zeta).
$$
This region contains covariate states that satisfy all rule constraints
not involving $X^{(\nu)}$, but are excluded from the original terminal
node by the constraints on $X^{(\nu)}$. 
An interval with
$\Z_{i,r}\in r^{\mathrm{miss}}_\nu(\zeta)$ is therefore a local near
miss for the rule. Comparing these intervals with those inside
$r(\zeta)$ assesses the contribution of $X^{(\nu)}$ to
hazard exposure, since if $\nu$ is a signal coordinate, the hazard
exposure within $r(\zeta)$ should be different from
$r^{\mathrm{miss}}_\nu(\zeta)$.

To localize the comparison in time, let
$\mathcal W_k=(t_{k-1},t_k]$ be a window formed by adjacent points on
the evaluation grid $\tt$, and define the intervals active in that
window by
$$
\mathcal A_k
=
\left\{
(i,r):
(S_{i,r},T_{i,r}]\cap\mathcal W_k\ne\emptyset
\right\}.
$$
For each active interval, the working response is the log integrated
out-of-bag hazard over its overlap with the window,
\Eq
W_{i,r,k}
=
\log\left\{
\int_{(S_{i,r},T_{i,r}]\cap\mathcal W_k}
\hhat_i^{\mathrm{oob}}(u)\,du
\right\}.
\label{E:rhf-varpro-response}
\EndEq

For a rule $\zeta$ and coordinate $\nu$, define indicators for
intervals inside the rule region and its near-miss region:
$$
I_{i,r}^{\mathrm{in}}(\zeta)
=
1_{\{\Z_{i,r}\in r(\zeta)\}},
\qquad
I_{i,r}^{\mathrm{miss}}(\zeta;\nu)
=
1_{\{\Z_{i,r}\in r^{\mathrm{miss}}_\nu(\zeta)\}}.
$$
Within time window $\mathcal W_k$, the in-rule mean and
near-miss mean are
$$
\widehat\theta_k^{\mathrm{in}}(\zeta)
=
\frac{
\sum_{(i,r)\in\mathcal A_k}
I_{i,r}^{\mathrm{in}}(\zeta)W_{i,r,k}
}{
\sum_{(i,r)\in\mathcal A_k}
I_{i,r}^{\mathrm{in}}(\zeta)
},
\qquad
\widehat\theta_k^{\mathrm{miss}}(\zeta;\nu)
=
\frac{
\sum_{(i,r)\in\mathcal A_k}
I_{i,r}^{\mathrm{miss}}(\zeta;\nu)W_{i,r,k}
}{
\sum_{(i,r)\in\mathcal A_k}
I_{i,r}^{\mathrm{miss}}(\zeta;\nu)
},
$$
whenever the denominators are positive.  The rule-level
contribution when these values are defined is
$$
\D_{\nu k}(\zeta)
=
\left|
\widehat\theta_k^{\mathrm{in}}(\zeta)
-
\widehat\theta_k^{\mathrm{miss}}(\zeta;\nu)
\right|.
$$

Let $\mathcal R$ denote the sampled collection of terminal-node rules
used in the forest-level calculation. For coordinate $\nu$ and time
window $\mathcal W_k$, let $\mathcal R_{\nu k}$ contain the rules in
$\mathcal R$ that use $X^{(\nu)}$ and for which both the in-rule and
near-miss means are defined. The time-localized priority score is
the average rule-level contribution,
$$
\widehat{\mathrm{VP}}_\nu(t_k)
=
\frac{1}{|\mathcal R_{\nu k}|}
\sum_{\zeta\in\mathcal R_{\nu k}}
\D_{\nu k}(\zeta),
\qquad
\mathcal R_{\nu k}\ne\emptyset.
$$
The curve $t_k\mapsto\widehat{\mathrm{VP}}_\nu(t_k)$ is the
time-localized variable priority curve for coordinate $\nu$. Large values identify
time windows in which $X^{(\nu)}$ has a large local effect on
the log integrated hazard.

\section{Empirical evaluation}

We evaluate the method on synthetic data with time-varying covariates.  Event
times are generated from latent continuous-time covariate
trajectories and the observed data consist of longitudinal
measurements sampled from those trajectories.  We vary the number of
records per subject over $R=10,20,40$.  Larger $R$ produces observed
histories with finer temporal resolution.  We compare RHF against
several competing methods (Section~\ref{subsec:competing-methods}).
Methods that accept time-updated covariates use the observed
predictable process in start--stop form, as in
Section~\ref{sec:rhf-startstop}.  Methods designed for fixed covariate
vectors use the corresponding baseline or landmark representations.
Each prediction uses only the covariate information available up to
the time of prediction.  Features that summarize how a covariate
changes over time, such as slopes or rates of change, enter the
predictor set only when the observed covariate vector already contains
them.

Complete details of the latent and observed covariate processes,
event-time and censoring mechanisms, simulation mechanisms, software
implementations, and tuning parameters for methods are given in the
Supplement, along with figures of representative hazard curves.
All analyses of the proposed method use the R package
\texttt{randomForestRHF}~\citep{randomForestRHF}.

\subsection{Time-varying covariate simulations}

The benchmark analysis uses three simulations that differ in how
time-dependent covariates shape the hazard.

\Enumerate

\item
\textit{Smooth time variation.}
The first simulation has a smoothly varying hazard.  Baseline
covariates determine the overall risk level, while the time-dependent
component is a scalar signal whose subject-specific slope is
determined by baseline covariates.
This smooth, low-dimensional structure is favorable to Cox regression
with time-dependent covariates, although the match to the generating
hazard is not exact (Section~S1.1 of the Supplement).

\item
\textit{Covariate-dependent zero-rate windows and latent longitudinal
  activation.}  In the second simulation,
baseline covariates define an extended interval of
exactly zero hazard. Outside this window, the hazard in one covariate
region is modified by a subject-specific longitudinal activation
process whose latent coefficients are not supplied to the learner.
The design tests whether a method can recover when risk is absent,
when it returns, and how it depends on the time varying longitudinal
state.

\item
\textit{Localized longitudinal risk region.}  In the third simulation,
risk is elevated only
when the current values of two longitudinal covariates jointly fall
within a specific region. Neither covariate alone
identifies this high-risk state. Thus the design tests whether a method can
track when a time varying covariate path occupies a localized
nonlinear risk region.

\EndEnumerate

\vspace*{-5pt}
\subsection{Evaluation metric for time-varying discrimination}

The benchmark methods used in our experiments
(Section~\ref{subsec:competing-methods}) utilize longitudinal
information in different ways and provide different types of outputs.
Some estimate hazards or survival curves, whereas others return
relative-risk scores or landmark-specific risks. Because these
prediction outputs differ in both scale and interpretation, no single
metric provides a complete comparison. We therefore evaluate
performance from two complementary perspectives. The first is
rank-based and measures time-varying discrimination: whether a method
assigns higher risk to subjects who experience the event by, or at, a
given time. We quantify discrimination using cumulative and incident
time-dependent AUCs.  The second assesses how accurately a method
recovers the underlying hazard, which we quantify using mean squared
error.

For the discrimination summaries, we evaluate each method using
two real-valued markers for subject $i$ at time $t$.  The cumulative
marker, denoted $M_i(t)$, summarizes risk accumulated up to $t$.  The
instantaneous marker, denoted $m_i(t)$, summarizes risk at or near
$t$, usually through a hazard estimate or a quantity proportional to
one.  Larger marker values always indicate higher risk.  The
cumulative AUC is computed from $M_i(t)$ and evaluates ranking over the
interval up to $t$.  The incident AUC is computed from $m_i(t)$ and
evaluates ranking locally at $t$.  For both quantities, 
we summarize performance
by an integrated AUC, denoted $\mathrm{iAUC}$.

All benchmark calculations are carried out on the original simulation
time scale, using the evaluation grid $\tt=\{0=t_0<t_1<\cdots<t_K\}$.
In each Monte Carlo replicate, this grid is constructed from the
observed event times in the training sample and then thinned so that
consecutive retained times have stable empirical event support; the
exact grid-construction rule is given in Supplementary Section~S1.
The integrated summaries reported below are therefore event-supported
grid averages, evaluated at common training-sample defined times for
all methods.  This convention concentrates the benchmark on regions of
follow-up where risk rankings and hazard estimates are empirically
well supported, while avoiding unstable tail contributions from
intervals with few observed events.

\subsubsection{Cumulative AUC using cumulative markers}

The cumulative AUC evaluates how well a cumulative marker $M_i(t)$
ranks subjects whose event time is no later than $t$ above subjects who
remain event-free beyond $t$.  For a fixed $t$ we define the population
quantity using the uncensored event time $T^*$,
$$
\AUC_{\mathrm{cum}}(t)
=
\PP\bigl\{ M_i(t) > M_j(t)
\;\big|\;
T_i^* \le t,\ T_j^* > t \bigr\},
$$
following the time-dependent ROC framework of
\citet{heagerty2000,heagerty2005}.  Here $(i,j)$ denotes a randomly
selected pair with subject $i$ failing by $t$ and subject $j$ remaining
event-free beyond $t$.

For the benchmark summaries, we evaluate this curve on $\tt$
and report the grid-weighted average
$$
\mathrm{iAUC}_{\mathrm{cum}}
=
\frac{
\sum_{k=1}^K w_k^{\mathrm{cum}}\AUC_{\mathrm{cum}}(t_k)
}{
\sum_{k=1}^K w_k^{\mathrm{cum}}
}.
$$
The nonnegative weights $w_k^{\mathrm{cum}}$ are event-supported
grid weights that account for the number of usable comparisons and
right censoring.  The finite-sample construction, including the
Kaplan--Meier estimator of the censoring distribution used in the
weights, is given in the Supplement (Section~S2).

\subsubsection{Incident AUC using instantaneous markers}

The incident AUC is the local-in-time analogue of the cumulative AUC.
At time $t$, it compares subjects who fail at that time with subjects
who are still at risk, using the instantaneous marker $m_i(t)$.  The
continuous-time target is
$$
\AUC_{\mathrm{inc}}(t)
=
\PP\bigl\{ m_i(t) > m_j(t)
\;\big|\;
T_i^*\in[t,t+dt),\ T_j^*> t \bigr\},
$$
again following \citet{heagerty2000,heagerty2005}.  Over the
evaluation grid this becomes
$$
\AUC_{\mathrm{inc}}(t_k)
=
\PP\bigl\{ m_i(t_k) > m_j(t_k)
\;\big|\;
T_i^*=t_k,\ T_j^*>t_k \bigr\}.
$$
The case group consists of subjects failing at the evaluation time,
and the control group consists of subjects who remain at risk at that
time.

In the benchmark, the reported incident summary is the
grid-weighted average
$$
\mathrm{iAUC}_{\mathrm{inc}}
=
\frac{
\sum_{k=1}^K w_k^{\mathrm{inc}}\AUC_{\mathrm{inc}}(t_k)
}{
\sum_{k=1}^K w_k^{\mathrm{inc}}
}.
$$
The weights $w_k^{\mathrm{inc}}$ play the same role as in the
cumulative AUC summary, giving more influence to grid times with
stable incident case--control comparisons while accounting for
right censoring.  The finite-sample approximation is described in the
Supplement.

\subsubsection{Relative hazard mean squared error}

Because the AUC summaries are rank-based, they do not assess whether a
method estimates the hazard on the correct numerical scale.  To
evaluate this directly, we define the relative mean squared error (rMSE)
for an estimator $\hhat$ by
$$
\frac{
\E\left[
\sum_{k=1}^K
\{\hhat(t_k,\X(t_k))-\haz(t_k,\X(t_k))\}^2
\right]
}{
\E\left[
\sum_{k=1}^K
\haz^2(t_k,\X(t_k))
\right]
}.
$$
The expectation is over an independent test subject generated from the
same simulation mechanism.  The denominator normalizes by the squared
true hazard, yielding a dimensionless measure for comparison across
simulation settings, with smaller rMSE indicating more accurate hazard
estimation.  The finite-sample approximation is given in the
Supplement.

\subsection{Competing methods}
\label{subsec:competing-methods}

The benchmark includes six competing methods.  Definitions, software
implementations, and marker constructions for all procedures are
provided in the Supplement.  Two of the methods use the same
start--stop representation as RHF and therefore have direct access to
the time-updated covariate process.  The \emph{Cox model with
time-dependent covariates} (Cox-TDC)~\citep{andersen1982cox} models
the hazard as $\haz_0(t)\exp\{\bbeta^T \Z_i(t)\}$.  Its instantaneous
marker is $m_i(t)=\exp\{\widehat{\bbeta}^T \Z_i(t)\}$, and its
cumulative marker is the cumulative hazard along the
start--stop path, $M_i(t)=\int_0^t\exp\{\widehat{\bbeta}^T
\Z_i(s)\}\,d\Hhat_0(s)$.  The \emph{boosted nonparametric hazard
estimator} (BoXHED) of \citet{lee2021boosted} also uses the
start--stop covariate process, but estimates the hazard
nonparametrically by gradient boosting under a hazard-likelihood loss.
For BoXHED \citep{pmlr-v119-wang20o, JSSv113i03}, the instantaneous
marker is the estimated case-specific hazard, and the cumulative
marker is obtained by integrating the hazard along the supplied
covariate path.

The remaining four methods are baseline and landmark-based survival
competitors.  \emph{Baseline RSF} fits a standard random survival
forest~\citep{ishwaran2008rsf}.  We use its estimated cumulative
hazard $\Hhat^{\mathrm{RSF}}(t\mid \X_i)$ as the cumulative marker and
finite-differenced increments of this estimate as the instantaneous
marker.  The \emph{discrete-time neural hazard model} (LogHazNN)
targets a piecewise-constant hazard using a logistic link and a fully
connected neural network
\citep{biganzoli1998feed,gensheimer2019scalable}.  \emph{DeepSurv}
replaces the linear predictor in the Cox model with a neural network
\citep{faraggi1995neural,katzman2018deepsurv}.  We fit both neural
baselines with the \texttt{pycox} library
\citep{kvamme2019time,kvamme2021continuous}.  The final method,
\emph{landmark random survival forest} (RSF-LM), is implemented as a
dynamic multi-landmark procedure
\citep{pickett2021random,moradian2022dynamic}.  At each landmark time
$\ell$, RSF is fit to subjects still at risk at $\ell$ using the
landmark covariates $\Z_i(\ell)$; predictions are obtained by
stitching increments from the most recent available landmark-specific
cumulative hazard.

\subsection{Benchmark design}
\label{subsec:benchmark-design}

\begin{figure}[phtb]
\centering
\vskip-5pt
\resizebox{5.25in}{!}{\includegraphics[page=1]{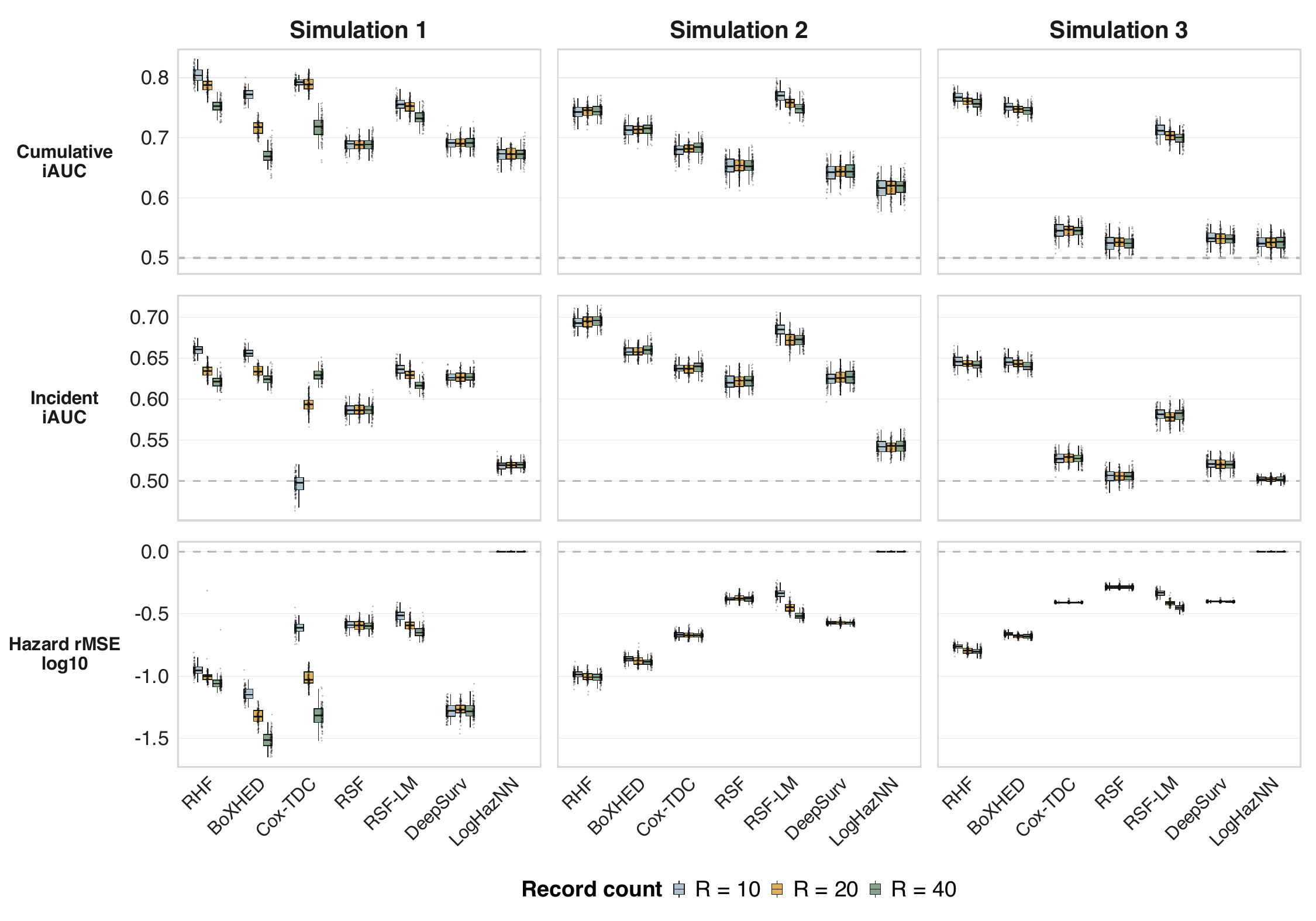}}
\vskip-10pt

\caption{\it Benchmark performance from time-varying covariate
  simulations.  Higher values indicate better
  discrimination for cumulative and incident $\mathrm{iAUC}$.  Lower
  values indicate more accurate hazard estimation for relative hazard
  MSE (displayed on $\log_{10}$ scale).  Dashed reference lines denote
  $\mathrm{iAUC}=0.5$ and $\log_{10}(\mathrm{rMSE})=0$.
  For readability, values of $\log_{10}(\mathrm{rMSE})$ outside
  $[-2,0]$ are clipped to the nearest boundary before constructing
  the boxplots.  Numerical summaries and rankings use the untruncated
  values.}
\label{benchmark}
\vskip5pt
\resizebox{4.25in}{!}{\includegraphics[page=1]{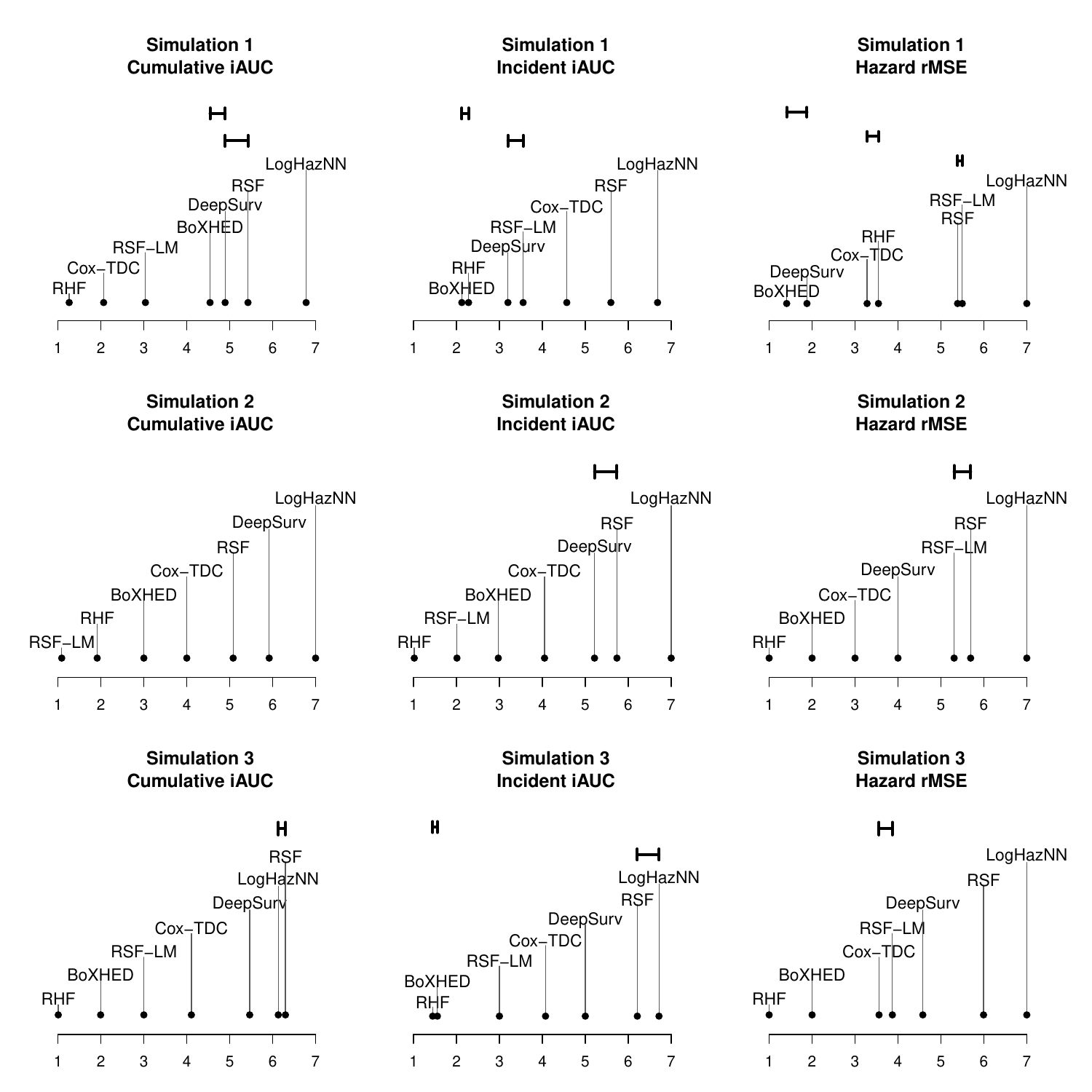}}
\vskip-5pt
\caption{\it Critical-difference plots for the benchmark results in
  Figure~\ref{benchmark}.  Points display average rank.
Horizontal
bars connect methods whose average ranks are within
the Nemenyi critical difference $\a=.01$.}
\label{benchmark.cd}
\vskip-5pt
\end{figure}

For each experiment we ran 100 independent Monte Carlo replicates.  In
each replicate, we drew independent training and test samples of equal
size $n=2500$ from the same data-generating mechanism.  Each method
was fit using only the training sample, and all time-dependent AUC
curves and performance summaries were computed from predictions on the
independent test sample.  Results from all experiments are displayed
in Figure~\ref{benchmark}, with critical-difference summaries given in
Figure~\ref{benchmark.cd}.

\subsection{Benchmark results}
\label{subsec:benchmark-results}

Simulation~1 provides a smooth, low-dimensional setting favorable to
regression-based methods. The baseline covariates determine both the
overall risk level and the slope of the longitudinal signal, so
methods using baseline information alone retain the variables that
drive the hazard. Cox-TDC achieves the highest cumulative
discrimination, while DeepSurv has the lowest hazard rMSE. Cox-TDC's
hazard rMSE also decreases substantially as $R$ increases from 10 to
40. This improvement is consistent with the finer temporal
representation of the longitudinal signal, as more frequent records
allow the time-updated covariates to track the underlying trajectory
more closely.

Simulation~2 presents a different challenge.  Risk can switch off over
a covariate-dependent time window and then reactivate through a
longitudinal signal whose latent coefficients are not supplied to the
learner.  This local on-off time structure is difficult to capture
with fixed-baseline summaries or simple log-linear effects.  In this
setting, RHF has the best overall performance, combining the lowest
hazard rMSE with the highest incident discrimination and near-highest
cumulative discrimination.  BoXHED is second in terms of hazard
estimation, while landmark RSF has the highest cumulative
discrimination but larger hazard rMSE.  The fixed-baseline forest and the neural methods
are the least accurate here.

In Simulation~3 we observe a clear separation between methods using
time-varying covariates from baseline methods.  This simulation is
especially challenging because risk is localized in the joint space of
two longitudinal coordinates.  RHF and BoXHED, which use time-varying
covariates effectively, are the top two in terms of cumulative and
incident discrimination. RHF has the lowest hazard rMSE.
Discrimination of the fixed-baseline methods is about the same as
random guessing.  Landmark RSF is better but not at the level of RHF
or BoXHED.

\subsection{Overall summary of benchmark results}
\label{subsec:overall-benchmark-results}

RHF ranks first in terms of rMSE and first or second in terms of
discrimination in the challenging Simulations~2 and~3. In
Simulation~1, a smooth, low-dimensional setting, it ranks between
first and third across the metrics. To assess whether these
conclusions also hold with smaller samples, we repeated all three
simulation experiments with $n=500$, retaining the same
data-generating mechanisms and nominal record counts $R=10,20,40$.
The results remain generally consistent with those at $n=2500$, with
RHF continuing to combine accurate hazard estimation with high
discrimination in the more challenging settings. These additional
experiments are reported in Section~S4 of the Supplement, with
the results displayed in Figure~S2.

\section{Time-varying ICU risk in MIMIC-IV}

We next apply the method to longitudinal ICU records from the Medical
Information Mart for Intensive Care IV (MIMIC-IV), a large, public
database of hospital and ICU encounters
\citep{Johnson2023mimiciv,Johnson2023mimicivicu}.  The analysis uses
the first ICU stay for each adult patient, age $\ge 18$ years, and
assembles approximately 11 million hourly records from routine
clinical care.  Predictors include physiologic measurements,
laboratory values, indicators of organ-support therapies, and
descriptors recorded at ICU admission.  This setting is a natural test
case for pathwise risk prediction because the information available
for prediction changes throughout the stay.  Measurements arrive
irregularly, different variables update on different schedules,
treatments begin and end, and patient condition can change rapidly.  A
baseline-only risk score compresses this into a single initial
summary, whereas the pathwise hazard trajectory updates as new
covariate values become available.  The case study has two goals.  We
first evaluate how well the fitted forest discriminates the
cause-specific in-hospital death.  We then use the time-localized variable priority scores from
Section~\ref{sec:rhf-vimp} to examine how variables drive risk
over time.

\subsection{Cohort construction}

The analysis targets the cause-specific hazard of in-hospital death
among patients who remain alive and hospitalized.  Discharge alive is
therefore treated as exit from the hospital-at-risk process, not as
independent censoring for an absolute death-before-discharge target.
Estimating absolute death-before-discharge probabilities would require
a competing-risk or multistate extension.  For subject $i$, let $T_i$
denote the time from ICU admission to the end of hospital-at-risk
follow-up, and let $\delta_i\in\{0,1\}$ denote the terminal status,
with $\delta_i=1$ for in-hospital death and $\delta_i=0$ for discharge
alive.

We restricted the analysis to the first ICU stay for each 
patient to avoid dependence across multiple stays
for the same person.  Follow-up begins at ICU admission and ends at the
earlier of in-hospital death or discharge alive, so there is at most one
event per stay.  After excluding stays with zero recorded duration, the
cohort contained $52{,}219$ patients, of whom $5{,}455$ died before
hospital discharge.  The start--stop dataset contained
$13{,}869{,}748$ hourly records.

Covariates were organized on a patient-specific grid of hourly risk
intervals anchored at ICU admission.  The grid aligned irregularly
recorded measurements and treatments with these intervals.  It did not
assume covariates were observed on a fixed hourly schedule.  Each row
held the information available at the start of its interval.  To
preserve temporal order, a measurement charted during an hour was
summarized within that hour but became available only at the following
interval.  Fixed lookback windows initialized only the admission-time
covariate history.  We included laboratory values charted within 48
hours before admission, and ICU-charted measurements and urine-output
records charted within 2 hours.  We filled short gaps after admission
by carrying the last observation forward, never backward, using a
6-hour window for vital signs and respiratory settings and a 24-hour
window for laboratory values.  We used no within-stay median
imputation.  When no prior value fell within the carry-forward
window, we recorded the value as missing and imputed a fixed,
predictor-specific value, the median estimated from the training
subjects only, which we applied unchanged to the testing cohort.  No
test-set information entered the preprocessing.  Missingness
indicators were not included as predictors in the primary
analysis.  Supplementary Section~S5 (Table~S2) summarizes
predictor-specific missingness, reporting counts and percentages of
hourly start--stop records with missing values after hourly feature
construction and before the carry-forward and fixed-value imputation
steps, and Supplementary Section~S6 reports a sensitivity analysis
that adds pre-imputation missingness indicators to the predictor set.
In addition, organ-support indicators were coded from the therapy
status available at the start of each interval.  Death and
discharge times defined only the start--stop intervals, the event
indicator, and the hospital-at-risk process.  We excluded discharge
disposition, length of stay, and other administrative outcome and
follow-up fields from the predictor matrix.  Taken together, these
preprocessing steps yielded a predictable hourly covariate process in
which patient-specific measurements and treatment states entered only
after they were available before the corresponding risk interval.

\subsection{Training and tuning parameters}

We split the cohort at the patient level, using $70\%$ of patients for
training and $30\%$ for testing, and kept each patient's full ICU
trajectory within a single split. We grew forests on the training
set using the same fixed tuning conventions as in the simulation
benchmarks, with $B=500$ trees, $\texttt{mtry}=\lceil\sqrt p\rceil$,
and an event-supported evaluation grid constructed from the training
sample as described in Supplementary Section~S1.

The remaining tuning parameter was tree size, the maximum number of
splits per tree, which we selected by minimizing the OOB empirical
risk described in Section~\ref{sec:rhf-oob}. For each
training patient, OOB predictions use only trees grown without any
of that patient's data. We evaluated candidate tree sizes from 5 to
100 in increments of 5 and selected tree size 90, which minimized
OOB risk (Figure~\ref{mimic_rhf_auc}, top panel). Selection used only
OOB risk; the test sample was reserved for performance evaluation.
Unless otherwise stated, all subsequent MIMIC-IV analyses use the
forest trained with this selected tree size.

\subsection{Performance evaluation}

We assessed hazard estimation in the held-out test sample using the
same likelihood loss as the OOB empirical risk. For each candidate
tree size, we evaluated this loss along the test patients' observed
covariate trajectories using predictions from the corresponding
training forest, and averaged the contributions over test patients.
The loss depends on the numerical magnitudes of the estimated hazards
through the at-risk exposure and observed-event contributions,
providing a direct scale-sensitive assessment that complements the
AUC measures of discrimination. Lower empirical risk indicates better
performance under this loss.

\begin{figure}[phtb]
  \centering
  \resizebox{3.5in}{!}{\includegraphics[page=1]{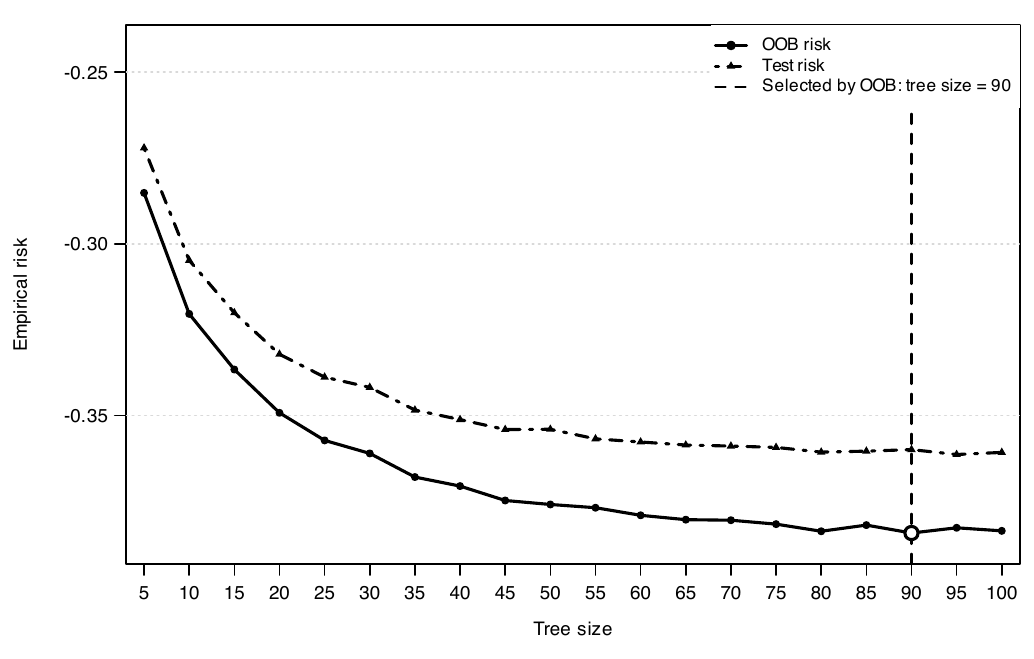}}
  \resizebox{4.0in}{!}{\includegraphics[page=1]{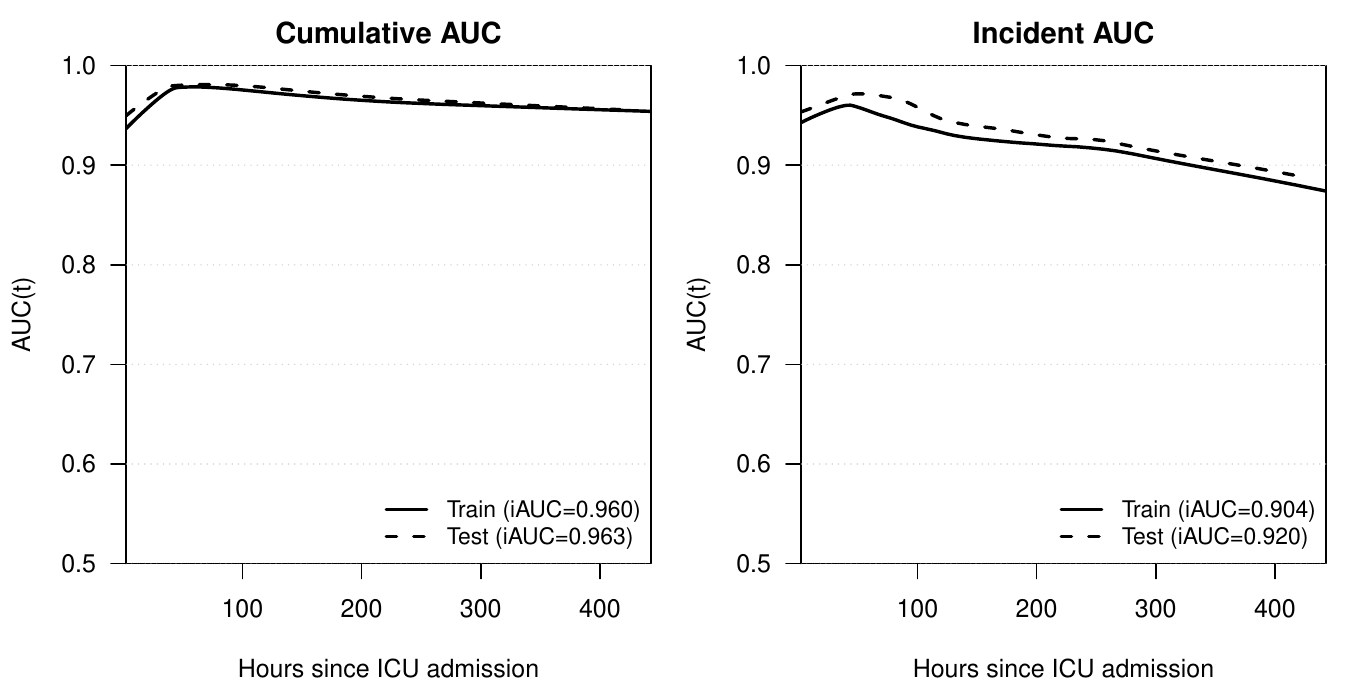}}
  \vskip-5pt
  \caption{\it Top panel shows OOB empirical risk (circles, solid line)
    and held-out test empirical risk (triangles, dashed line) versus tree size,
    computed using the likelihood loss in
    Section~\ref{sec:rhf-oob}; lower values are better. The vertical
    dashed line marks tree size 90, selected solely by minimizing
    OOB risk. Test risk is shown for evaluation and was not used for
    tuning. Bottom panels show cumulative (left) and incident
    (right) AUC curves for the selected forest, using OOB predictions
    in the training set (solid lines) and held-out test predictions
    (dashed lines). AUC follow-up time is in hours since ICU
    admission.}
  \label{mimic_rhf_auc}
  \vskip-5pt
\end{figure}

The top panel of Figure~\ref{mimic_rhf_auc} compares OOB and test
risk across candidate tree sizes. Both curves decrease substantially
before leveling off at larger tree sizes. Although test risk is
consistently higher than OOB risk, it follows a similar pattern,
showing that the improvements identified by OOB tuning extend to
held-out patients. The selected tree size lies in the region where
test risk is nearly flat. This agreement in the tree-size pattern
provides an additional empirical check on the use of OOB risk for
tuning.

The likelihood loss evaluates the estimated hazards on their numerical
scale along the observed trajectories, but it is not a calibration
diagnostic.  We did not assess calibration of the estimated hazard
trajectories, for example by comparing expected and observed event
counts within risk strata over follow-up.  The MIMIC-IV results
therefore establish discrimination and hazard-scale fit under the
likelihood loss, and calibration issues are left for future work
(Section~7).

For the trained forest, we also evaluated the cumulative and
incident AUC curves defined in Section~5, using OOB predictions for
the training sample and full-forest predictions for the test sample.
The bottom panels of Figure~\ref{mimic_rhf_auc} show high
discrimination of the cause-specific in-hospital death hazard in
both samples. Cumulative AUC is in the mid $0.9$ range across most
of follow-up, while incident AUC, a more local measure, is in the
low $0.9$ range.

\subsection{Time-localized variable priority for MIMIC-IV}
\label{sec:mimic-varpro}

\begin{figure}[phtb]
  \centering
  \resizebox{5.5in}{!}{\includegraphics[page=1]{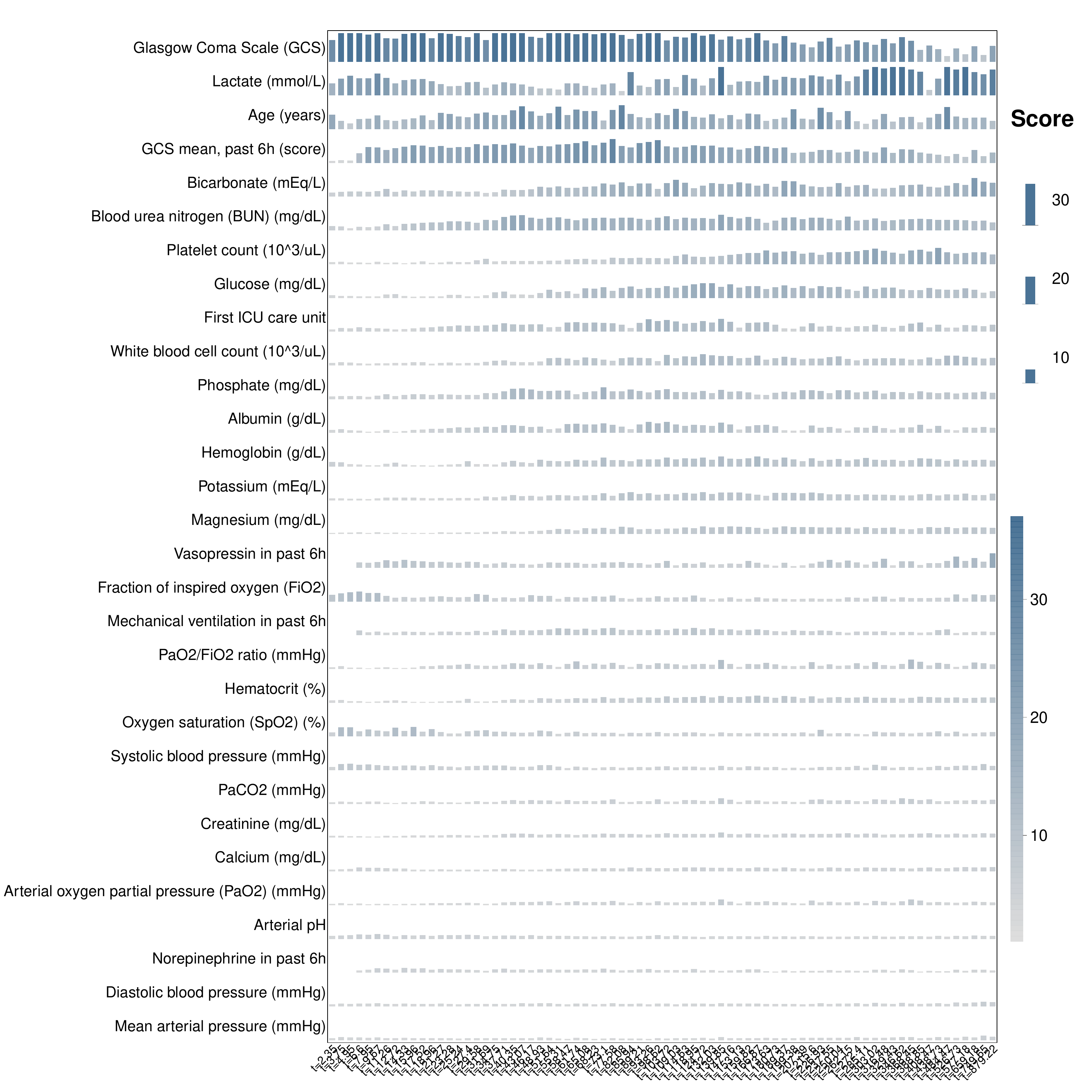}}
  \vskip-5pt
  \caption{\it Time-localized variable priority analysis for
  MIMIC-IV.  For each evaluation-grid window, the calculation is
  restricted to start--stop intervals that overlap the window and
  compares the log OOB integrated hazard exposure inside each forest
  rule with that of the corresponding near-miss set obtained by
  releasing a variable.  Each bar shows the priority score
  for a variable at a given time.  Bar size and color increase with
  the priority score, with legends shown at right.}
  \label{mimic_rhf_vimp}
\end{figure}

To better understand which measurements drive the hazard at
different stages of an ICU stay, we applied the time-localized variable
priority method of Section~\ref{sec:rhf-vimp}.  The calculation uses the
same start--stop representation used to fit the forest.  Each interval was
assigned the log OOB integrated hazard exposure
in~\eqref{E:rhf-varpro-response}, and, for each evaluation-grid window,
the rule and near-miss summaries were computed using only intervals that
overlapped that window.  Thus, the score for a given window reflects
local rule contrasts among the interval records contributing to the
hospital-at-risk process during that period.  Figure~\ref{mimic_rhf_vimp}
displays the resulting variable priority scores as a barplot matrix.  Time
windows are shown on the horizontal axis, variables on the vertical
axis, and bar size and color encode the magnitude of the priority
score.

Neurologic status is a key contributor to the profile plot. In
particular, the Glasgow Coma Scale has large scores throughout; its
6-hour mean is especially large early on. Lactate is also a top
variable, with scores that increase over time. Age, blood urea
nitrogen, and bicarbonate have large scores throughout, whereas
platelet count, white blood cell count, glucose, hemoglobin,
phosphate, potassium, and other laboratory markers have larger scores
later. Vasopressin use also appears at several time points.
Oxygenation measures, including oxygen saturation, fraction of
inspired oxygen, and the $\mathrm{PaO}_2/\mathrm{FiO}_2$ ratio, have
more localized and generally smaller scores. Overall, the priority
analysis points to a temporal shift in the hazard, in which neurologic
status and acute metabolic stress are early indicators, while renal,
hematologic, metabolic, and treatment-related markers play a role in
longer stays.

\section{Discussion}

This paper introduced Random Hazard Forests (RHF), a survival tree
ensemble that estimates a hazard map on time--covariate space for
time-dependent covariates. Applying this map to a
predictable covariate process yields a path-specific hazard trajectory
that evolves as the covariate state changes. Because each state is
determined by information available immediately beforehand, the
construction preserves temporal ordering and accommodates internal
longitudinal covariates without lookahead. The method thereby models evolving
risk directly, without reducing the problem to baseline covariates,
landmark snapshots, or fixed discrete-time classification problems.

The experiments clarify when this type of flexibility is most helpful
and when it is not.  In Simulation~1, where the hazard was smooth and
low dimensional, regression-based methods, such as Cox-TDC, matched or
outperformed RHF on some metrics.  On the other hand, in Simulations~2
and~3, where risk depended on more complex features of the current
longitudinal state, the method was among the best in terms of
discrimination and ranked first for hazard rMSE.  This flexibility
therefore becomes more valuable as the complexity of risk increases.

The MIMIC-IV case study illustrated an additional advantage of the
pathwise formulation. Alongside its predictive performance, RHF
revealed how the variables contributing to the hazard changed over
follow-up. The time-localized variable priority analysis distinguished
predictors whose contributions persisted throughout follow-up from
those whose contributions were concentrated early or late. Clinical
predictors were the top variables even when pre-imputation missingness
indicators were included (Supplementary Section~S6).  The
highest-ranked indicator placed 21st, well below the top clinical
predictors.  By resolving predictor relevance over time, the analysis
revealed patterns that a single global ranking would miss. This
perspective is especially useful in clinical settings, where
physiology and treatment evolve over time and their changing relevance
can inform patient management.

The present work also leaves several directions open. The
current formulation targets a single cause-specific event process, but
the same likelihood-based ideas could be developed for competing
risks, recurrent events, and multistate outcomes. Because discrimination
and hazard accuracy do not by themselves ensure well-calibrated
estimates, calibration diagnostics and post-hoc calibration methods
for hazard trajectories are also needed.  The MIMIC-IV analysis
reported discrimination and likelihood-based hazard fit but did not
assess calibration of the estimated trajectories.  For routine-care data, the
observation process is itself part of the problem: missingness,
irregular measurement, and informative observation times should be
modeled more explicitly. In the MIMIC-IV analysis, we did not evaluate
sensitivity to alternative carry-forward windows or imputation rules.
Streaming implementations and richer local explanation tools,
including individual-level variable priority \citep{lu2025individual},
would further support real-time clinical use.

\begin{funding}
Hemant Ishwaran and Udaya B.~Kogalur were supported by the National
Institutes of Health through the National Institute of General Medical
Sciences (R35 GM139659) and the National Heart, Lung, and Blood
Institute (R01 HL164405).  Eileen M.~Hsich and Donald K.~K.~Lee were
supported by the National Institutes of Health through the National
Heart, Lung, and Blood Institute (R01 HL164405).
\end{funding}

\section*{Data and Code Availability}
The \texttt{randomForestRHF} R-package implementing Random Hazard
Forests
is publicly available on CRAN~\citep{randomForestRHF}.  Functions for
implementing simulations are included in the package, and detailed
descriptions of the benchmark experiment are provided in the
Supplement for reproducibility.  The MIMIC-IV data used in the ICU
case study are publicly available through PhysioNet
\citep{Johnson2023mimiciv,Johnson2023mimicivicu} after completing the
required credentialing process.

\vspace*{20pt}

\bibliographystyle{imsart-nameyear}
\bibliography{references}

\end{document}


\begin{frontmatter}
\title{Supplement to ``Random Hazard Forests''}
\runtitle{Supplement to Random Hazard Forests}

\begin{aug}
\author[A]{\fnms{Hemant}~\snm{Ishwaran}\ead[label=e1]{hishwaran@miami.edu}\orcid{0000-0003-2758-9647}},
\author[B]{\fnms{Eileen M.}~\snm{Hsich}\ead[label=e2]{eileen.hsich@imail.org}},\\
\author[C]{\fnms{Udaya B.}~\snm{Kogalur}\ead[label=e3]{ubkogalur@gmail.com}}
\and
\author[D]{\fnms{Donald K.~K.}~\snm{Lee}\ead[label=e4]{donald.lee@emory.edu}}

\address[A]{Division of Biostatistics,
Department of Public Health Sciences,
Miller School of Medicine,
University of Miami\printead[presep={,\ }]{e1}}

\address[B]{Heart and Vascular Institute,
Cleveland Clinic; Cardiology Division,
Intermountain Health%
\printead[presep={,\ }]{e2}}

\address[C]{Kogalur \& Company, Inc.\printead[presep={,\ }]{e3}}

\address[D]{%
  Goizueta Business School
  and Department of Biostatistics \& Bioinformatics, 
  Emory University\printead[presep={,\ }]{e4}}
\end{aug}

\end{frontmatter}

\beginsupplement

This supplement provides additional details on the simulation designs,
observed covariate processes, and definitions of the risk markers and
hazard-based accuracy measures used in the empirical benchmark analysis
in the main paper.  It reports the benchmark results for the smaller
training sample size $n=500$.  It also summarizes predictor
missingness in the MIMIC-IV case study and reports a sensitivity
analysis that includes missingness indicators.

\section{Simulation details}

This section gives the complete data-generating mechanisms for the three
time-varying covariate simulations in the main paper.  Each simulation
uses a latent continuous-time process, denoted by
$\widetilde{\X}_i(t)$, to generate the event-time hazard.  The covariate
process available to the learning methods is the observed predictable
process $\X_i(t)$, represented in start--stop form by row-level vectors
$\Z_{i,r}$.  Specifically, follow-up for subject $i$ is partitioned into
intervals $(S_{i,r},T_{i,r}]$, and $\Z_{i,r}$ denotes the covariate state
associated with interval $r$, $r=1,\ldots,R_i$.  These row-level vectors
define the observed predictable process
$$
\X_i(t)=\Z_i(t)
=
\sum_{r=1}^{R_i}\Z_{i,r}
1_{\{t\in(S_{i,r},T_{i,r}]\}}.
$$
Thus $\widetilde{\X}_i(\cdot)$ denotes the latent trajectory entering the
event-time hazard, whereas $\X_i(\cdot)$, equivalently $\Z_i(\cdot):=\{\Z_{i,r}\}$, denotes the predictable covariate
process represented by the start--stop data.

For each subject $i$, baseline covariates are generated as
$$
\widetilde{\X}_i(0)=(X_i^{(1)},\ldots,X_i^{(p)})^T,
\qquad X_i^{(k)}\iid U[0,1],
$$
unless a simulation states otherwise; the numerical experiments in the
main paper use $p=10$.  Only the coordinates explicitly listed below
affect the hazard; additional coordinates are independent noise
variables.  Conditional on the latent path, the event time $T_i^*$ is
generated from the continuous-time hazard
$\haz(t,\widetilde{\X}_i(t))$.  Equivalently, if
$E_i\sim\mathrm{Exp}(1)$, then
$$
T_i^*=\inf\left\{t>0:
\int_0^t \haz(s,\widetilde{\X}_i(s))\,ds\ge E_i\right\}.
$$
Independent right censoring is then applied by generating a censoring
time $C_i$ independently of the event process and setting
$$
T_i=\min\{T_i^*,C_i\},
\qquad
\delta_i=1_{\{T_i^*\le C_i\}}.
$$
The same observed training and test data are used for all methods within
a Monte Carlo replicate.  The censoring distribution differs by
simulation and is specified below.

To mimic irregular longitudinal sampling, let $R>1$ denote the nominal
record-count parameter.  The realized number of records for subject $i$
is
$$
B_i\sim\operatorname{Binomial}(R-1,0.7),
\qquad R_i=1+B_i.
$$
After $T_i$ has been generated, we draw
$R_i-1$ internal stop times independently from $U(0,T_i)$, sort them,
and append the observed endpoint $T_i$.  This gives contiguous intervals
with
$$
0=S_{i,1}<T_{i,1}=S_{i,2}<\cdots<T_{i,R_i-1}=S_{i,R_i}<T_{i,R_i}=T_i.
$$
The main paper reports results for $n=2500$ and nominal values
$R=10,20,40$.  Results for $n=500$ are reported in
Section~\ref{subsec:supp-benchmark-n500}.  For each simulation scenario,
each value of $R$, and each sample size shown in the figures, we use
100 independent Monte Carlo replicates.
In every replicate, the training and test samples are generated
independently from the same mechanism and have the same sample size.

For each time-dependent coordinate, the value used on a start--stop
record is the value of the corresponding trajectory on that interval.
In the notation above, if $\psi_i(t)$ is a scalar time-dependent
trajectory, then the row-level covariate is
$$
Z_{i,r}^{\mathrm{tdc}}=\psi_i(T_{i,r}),
\qquad r=1,\ldots,R_i.
$$
For the continuous trajectories considered here, this is interpreted
as the covariate value immediately before the right endpoint of the
interval, so the resulting row construction agrees with the
predictable start--stop convention.  Equivalently,
$Z_{i,r}^{\mathrm{tdc}}$ defines the corresponding coordinate of
$\X_i(t)$ on $(S_{i,r},T_{i,r}]$.  Static baseline coordinates are
repeated on every row.

No derivative, slope, or change-rate feature is added unless it is
explicitly included in the fitted covariate vector.  Thus, in
Simulations~2 and~3, the latent intercepts and slopes used to generate
the longitudinal trajectories are not supplied directly to the learner.
They can only be inferred indirectly from the repeated observed
measurements.

All formulas below are written on the original simulation time scale.
In each Monte Carlo replicate, the common evaluation grid is
constructed from the observed event times in the training sample.  Let
$\tt=\{0=t_0<t_1<\cdots<t_K\}$ denote this grid.  The grid is chosen
so that each interval $I_k=(t_{k-1},t_k]$ contains at least $m$
  training events, subject to the upper bound $K\le K_{\max}$ on the
  number of grid intervals.  All methods are evaluated on this same
  grid, and all benchmark summaries in this supplement are reported on
  the original simulation time scale.  All simulation analyses used
  $K_{\max}=100$ and $m=50$.

\subsection{Simulation 1: smoothly varying hazard}
\label{subsec:supp-sim1}

The first simulation is a smooth time-varying design.  All baseline
covariates are time-static.  The only time-updated signal is a scalar
linear trajectory,
$$
w_i(t)=\{X_i^{(4)}+X_i^{(5)}\}t,
$$
whose slope is determined by two baseline covariates.  For compactness,
let
$$
\zeta_i=1.5\{X_i^{(4)}+X_i^{(5)}\}.
$$
The event-time hazard is
$$
\haz(t,\widetilde{\X}_i(t))
=
(1+\zeta_i t)\exp(\zeta_i t)\exp(1.5X_i^{(1)}),
$$
and the cumulative hazard is
$$
\Haz_i(t)
=
\int_0^t \haz(s,\widetilde{\X}_i(s))\,ds
=
t\exp\{1.5X_i^{(1)}+\zeta_i t\}.
$$
Thus $T_i^*$ is obtained by solving $\Haz_i(T_i^*)=E_i$.  If
$\zeta_i=0$, this gives
$T_i^*=E_i\exp(-1.5X_i^{(1)})$; otherwise the equation is solved by
one-dimensional numerical inversion.  Independent censoring is
generated as
$$
C_i=-1.5\log(U_i^c),
\qquad U_i^c\sim U[0,1].
$$
The factor $\exp(1.5X_i^{(1)})$ controls the overall risk level, while
$(X_i^{(4)},X_i^{(5)})$ controls the smooth time trend through
$\zeta_i$.  This smooth, low-dimensional structure is favorable to
regression-based methods, including the Cox-TDC procedure used in the
simulations.

The observed start--stop data contain the baseline coordinates,
repeated across records, together with the sampled time-dependent
column
$$
Z_{i,r}^{\mathrm{tdc}}
=
w_i(T_{i,r})
=
\{X_i^{(4)}+X_i^{(5)}\}T_{i,r}.
$$
The generating log hazard is
$$
\log\{\haz(t,\widetilde{\X}_i(t))\}
=
1.5X_i^{(1)}+1.5w_i(t)+\log\{1+1.5w_i(t)\}.
$$
Note that Cox-TDC uses additive linear terms for the baseline
covariates and the sampled time-dependent signal.  It omits
$\log\{1+1.5w_i(t)\}$, which is nonlinear in the signal and varies
across subjects at a fixed time.  This term cannot be absorbed into a
common baseline hazard.  Thus, the linear Cox-TDC procedure is not
exactly specified, even if $w_i(t)$ were observed continuously.  The
simulation becomes more favorable to Cox-TDC as $R$ increases, but the
model never matches the generating hazard exactly.

\subsection{Simulation 2: covariate-dependent zero-rate windows and latent longitudinal activation}

The second simulation creates covariate-dependent zero-rate windows and a
longitudinal activation effect that is not determined by the baseline
covariates.  We generate 
subject-specific trajectory parameters
$$
A_i\sim N(0,0.8^2),
\qquad
B_i\sim N(0,0.45^2),
$$
truncated at four standard deviations from their means.  The latent
time-dependent coordinate is
$$
\psi_i(t)=A_i+B_i t.
$$
The parameters $(A_i,B_i)$ are not supplied as separate predictors; only
repeated observed values of the trajectory are available through the
start--stop data.

The hazard is selected by the baseline covariate region.  If
$X_i^{(1)}\le 0.5$ and $X_i^{(2)}\le 0.5$, then
$$
\haz(t,\widetilde{\X}_i(t))=
\haz_1(t,\widetilde{\X}_i(t))
=
\begin{cases}
0, & 0.5\le t\le 2.5,\\[4pt]
\exp(0.2X_i^{(2)}), & \text{otherwise}.
\end{cases}
$$
Otherwise,
$$
\haz(t,\widetilde{\X}_i(t))=
\haz_2(t,\widetilde{\X}_i(t))
=
\begin{cases}
0, & 2.5\le t\le 4.5,\\[4pt]
\exp\{0.2X_i^{(3)}+\gamma 1_{\{\psi_i(t)>0\}}\}, & \text{otherwise}
\end{cases}
$$
with $\gamma=1.1$.

For the first region, writing $a_i=\exp(0.2X_i^{(2)})$, the cumulative
hazard is
$$
\Haz_{1i}(t)
=
a_i\left\{\min(t,0.5)+(t-2.5)_+\right\},
$$
where $(u)_+=\max(u,0)$.  For the second region, let
$b_i=\exp(0.2X_i^{(3)})$ and define
$$
L_i(c,d)=\int_c^d 1_{\{\psi_i(s)>0\}}\,ds,
\qquad 0\le c<d,
$$
with $L_i(c,d)=0$ when $d\le c$.  Then
\begin{eqnarray*}
\Haz_{2i}(t)
&=&
b_i\Big[
\min(t,2.5)+(t-4.5)_+\\
&& \qquad +
\{\exp(\gamma)-1\}
\left\{
L_i(0,\min(t,2.5))
+
1_{\{t>4.5\}}L_i(4.5,t)
\right\}
\Big].
\end{eqnarray*}
The event time is obtained by solving
$\Haz_{1i}(T_i^*)=E_i$ or $\Haz_{2i}(T_i^*)=E_i$, depending on the
subject's covariate region.  Independent censoring is generated as
$$
C_i=-5.5\log(U_i^c),
\qquad U_i^c\sim U[0,1].
$$

The data include the baseline covariates together with the
row-level time-dependent coordinate
$$
Z_{i,r}^{\mathrm{tdc}}
=
\psi_i(T_{i,r})=A_i+B_iT_{i,r}.
$$
This design is difficult for
methods that collapse the longitudinal history to a baseline vector,
because the activation process $\psi_i(t)$ is not determined by the
baseline covariates.  It is also challenging for log-linear
relative-risk models because the activation effect is thresholded and
the zero-rate windows depend on the covariate region.

\subsection{Simulation 3: localized longitudinal risk region}

The third simulation creates a nonlinear risk effect that depends on the
current value of two longitudinal coordinates.
We independently generate
unobserved latent intercepts
$$
A_{4i},A_{5i}\iid U[0,1]
$$
and unobserved latent slopes
$$
B_{4i},B_{5i}\iid N(0,0.5^2),
$$
truncated at four standard deviations from their means.  The two latent
time-dependent longitudinal trajectories are
$$
\widetilde{X}_i^{(4)}(t)=A_{4i}+B_{4i}t,
\qquad
\widetilde{X}_i^{(5)}(t)=A_{5i}+B_{5i}t.
$$

Let
$$
r_i(t)
=
1_{\{0.25<\widetilde{X}_i^{(4)}(t)<0.85,
      \;0.20<\widetilde{X}_i^{(5)}(t)<0.80\}}
$$
denote the indicator that the subject's current longitudinal state lies
inside the local risk region.  Let $\gamma=1.5$ and
$(\beta_1,\beta_2)=(0.3,-0.3)$. The hazard is defined by
$$
\haz(t,\widetilde{\X}_i(t))
=
(1+t)
\exp\{\beta_1X_i^{(1)}+\beta_2X_i^{(2)}+\gamma r_i(t)\}.
$$
This is nonlinear in the time-dependent covariates,
with risk elevated in an interior rectangle of the longitudinal state
space.

The cumulative hazard has a simple piecewise form.  Let
$$
G(a,b)=\left(b+{b^2\over 2}\right)-\left(a+{a^2\over 2}\right),
\qquad 0\le a<b,
$$
and let $(L_i,U_i)$ be the interval of times for which the two
longitudinal trajectories are simultaneously inside the local risk
region.  If this interval is empty, define $J_i(t)=0$ for all $t$;
otherwise set
$$
J_i(t)=
\begin{cases}
0, & t\le L_i,\\[3pt]
G\Big(L_i,\min(t,U_i)\Big), & t>L_i.
\end{cases}
$$
Then
$$
\Haz_i(t)
=
\exp\{\beta_1X_i^{(1)}+\beta_2X_i^{(2)}\}
\left[
G(0,t)+\{\exp(\gamma)-1\}J_i(t)
\right].
$$
The event time is obtained by solving $\Haz_i(T_i^*)=E_i$ by
one-dimensional root finding.  Independent censoring is generated as
$$
C_i=-4\log(U_i^c),
\qquad U_i^c\sim U[0,1].
$$

The data include row-level values of the two longitudinal
coordinates:
$$
Z_{i,r}^{(4)}=A_{4i}+B_{4i}T_{i,r},
\qquad
Z_{i,r}^{(5)}=A_{5i}+B_{5i}T_{i,r}.
$$
Because the latent intercepts and slopes are not determined by the
baseline covariates, methods that collapse the trajectory to a baseline
vector lose information about the subject's current longitudinal risk
state.

\begin{figure}[phtb]
\centering
\includegraphics[width=\textwidth]{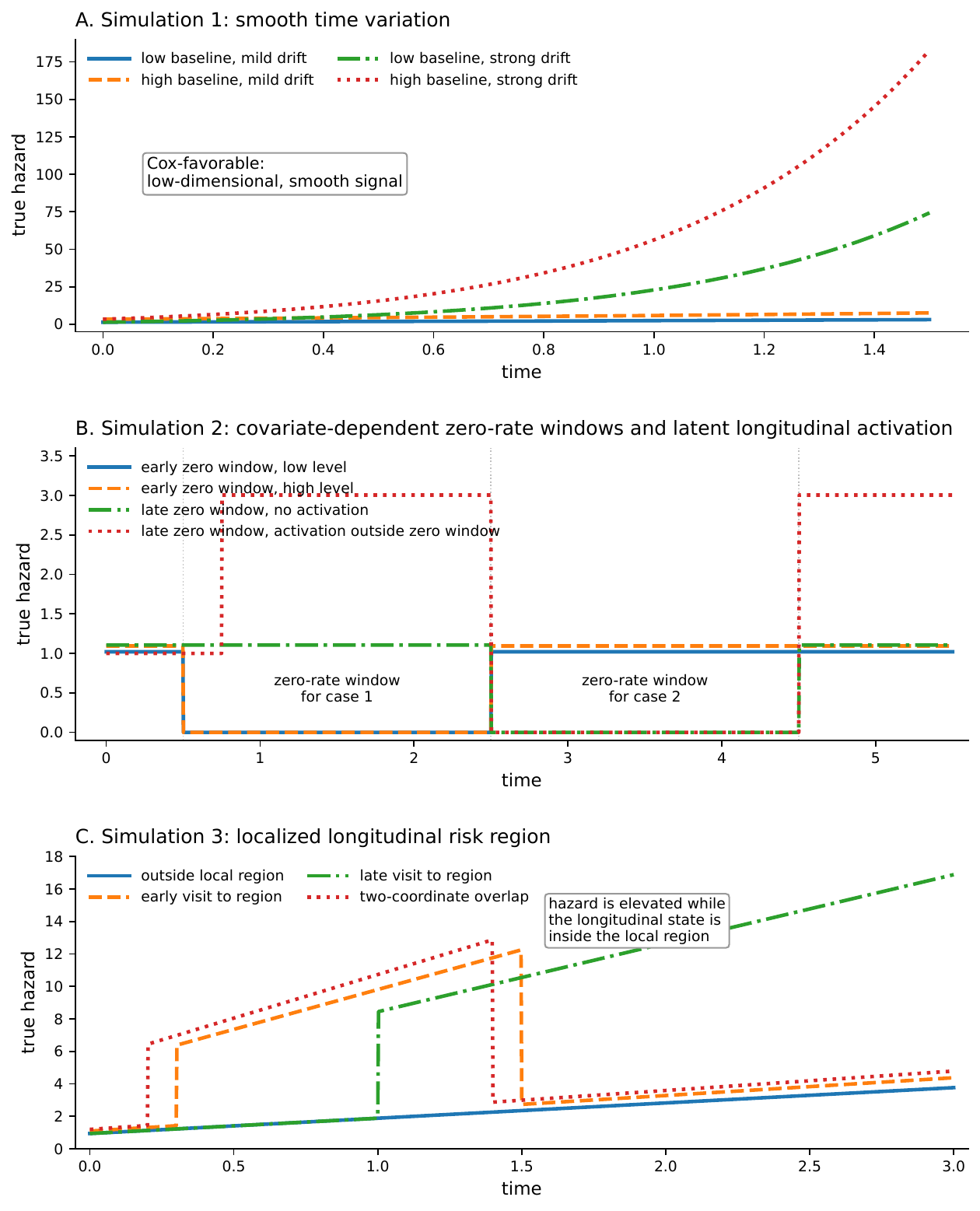}

\caption{Representative true hazard curves for the three simulation
  scenarios.  Panel~A shows smoothly varying hazards from
  Simulation~1, illustrating changes in baseline level and time trend.
  Panel~B shows the covariate-dependent zero-rate windows in
  Simulation~2.  The blue and orange curves represent the first
  covariate region, where the hazard is zero on the early window
  $[0.5,2.5]$ and otherwise constant at different levels.  The green
  and red curves represent the second covariate region, where the
  hazard is zero on the later window $[2.5,4.5]$; the green curve has
  no latent activation, whereas the red curve illustrates the jump in
  hazard that occurs when the latent longitudinal process crosses its
  activation threshold outside the zero-rate window.  Panel~C shows
  representative hazards from Simulation~3.  The curves illustrate
  subjects whose longitudinal trajectories spend different amounts of
  follow-up inside the local risk region, producing time-local
  increases in hazard when the current longitudinal state enters the
  interior rectangle. }

\label{fig:supp-hazard-shapes}
\end{figure}

\subsection{Illustrative hazard shapes}
Figure~\ref{fig:supp-hazard-shapes} plots representative true hazard
curves for the three simulation scenarios.
Panel~A shows Simulation~1.  The smooth, low-dimensional hazard is
favorable to Cox-TDC, although its linear predictor omits the nonlinear
term identified in Section~\ref{subsec:supp-sim1}.
Panel~B shows Simulation~2. The design combines covariate-dependent
zero-rate windows with a subject-specific longitudinal activation
effect that operates outside those windows. Two subjects can share
similar overall exposure or event rates while their risk activates at
sharply different times. Baseline covariates alone do not determine
this activation, so any method that collapses a subject to a single
baseline vector discards case-specific information about the current
risk state.
Panel~C shows Simulation~3. Risk rises when the current pair of
longitudinal coordinates lies inside an interior rectangle of the
longitudinal state space. A subject's hazard climbs as its trajectory
enters this region and falls back to baseline once the trajectory
leaves. Baseline-vector methods cannot track this movement, and a
log-linear time-dependent Cox model cannot capture the rectangular
effect.

\section{Procedures, AUC markers, and software implementations}

This section describes the finite-sample AUC estimators, software
implementations, tuning settings, and method-specific cumulative and
instantaneous markers used in the benchmark analysis.  The cumulative
marker $M_i(t)$ is used for the cumulative AUC, and the instantaneous
marker $m_i(t)$ is used for the incident AUC, as described in
Section~5.2 of the main paper.

\subsection{Finite sample AUC approximations}

We first describe the finite-sample approximation used to estimate the
AUC metrics.  Assume that the cumulative marker $M_i(t)$ and
instantaneous marker $m_i(t)$ are given.  Given markers $a$ and $b$, define
$$
\phi(a,b)=1_{\{a>b\}}+\frac{1}{2}1_{\{a=b\}},
$$
so that tied marker values receive one-half credit.

For the cumulative AUC at an evaluation time $t_k$, define the
case and control sets
$$
\mathcal C_k^{\mathrm{cum}}
=
\{i:\delta_i=1,\ T_i\le t_k\},
\qquad
\mathcal R_k
=
\{j:T_j>t_k\}.
$$
The time-specific empirical cumulative AUC is
$$
\widehat\AUC_{\mathrm{cum}}(t_k)
=
\frac{
\sum_{i\in\mathcal C_k^{\mathrm{cum}}}
\sum_{j\in\mathcal R_k}
\phi\{M_i(t_k),M_j(t_k)\}
}{
|\mathcal C_k^{\mathrm{cum}}|\,|\mathcal R_k|
}.
$$
For the incident AUC, events are assigned to the nearest evaluation grid
time on the right.  Equivalently, for
$I_k=(t_{k-1},t_k]$, define the case sets as
$$
\mathcal C_k^{\mathrm{inc}}
=
\{i:\delta_i=1,\ T_i\in I_k\}.
$$
The time-specific empirical
incident AUC is
$$
\widehat\AUC_{\mathrm{inc}}(t_k)
=
\frac{
\sum_{i\in\mathcal C_k^{\mathrm{inc}}}
\sum_{j\in\mathcal R_k}
\phi\{m_i(t_k),m_j(t_k)\}
}{
|\mathcal C_k^{\mathrm{inc}}|\,|\mathcal R_k|
}.
$$

The reported integrated AUC summaries are weighted averages over the
evaluation times with finite time-specific AUC estimates.  Let
$\widehat G_C$ denote the Kaplan--Meier estimator of the censoring
survival function.  For $a\in\{\mathrm{cum},\mathrm{inc}\}$, the
time weight used in the benchmark is
$$
w_k^a
=
\frac{|\mathcal C_k^a|}
     {\{\max(\widehat G_C(t_k),g_0)\}^{2}},
$$
where $g_0=0.10$ is the censoring-survival floor used in the benchmark.
The reported summary is
$$
\widehat{\mathrm{iAUC}}_{a}
=
\frac{\sum_k w_k^a\widehat\AUC_a(t_k)}
     {\sum_k w_k^a},
\qquad a\in\{\mathrm{cum},\mathrm{inc}\}.
$$

\subsection{Software and tuning overview}

Table~\ref{tab:benchmark-tuning} summarizes the data input
representation, software, and tuning settings used by each method.
Unless otherwise stated, these settings were fixed in advance and used
unchanged across simulation settings, sample sizes, record
frequencies, and Monte Carlo replicates.  Default package settings
were used for parameters not listed in the table.  The method-specific
subsections below describe the prediction conventions and define the
cumulative and instantaneous markers used in the AUC calculations.

\begingroup
\footnotesize
\setlength{\tabcolsep}{5pt}
\renewcommand{\arraystretch}{1.12}
\setlength{\LTcapwidth}{\textwidth}
\begin{longtable}{@{}p{0.14\textwidth}p{0.12\textwidth}p{0.20\textwidth}p{\dimexpr0.54\textwidth-6\tabcolsep\relax}@{}}
\caption{Input representations, software, and tuning settings used in
the benchmark simulations.}
\label{tab:benchmark-tuning}\\
\hline
Procedure & Input & Software & Settings \\
\hline
\endfirsthead
\multicolumn{4}{@{}l}{Table~\thetable\ (continued)}\\[3pt]
\hline
Procedure & Input & Software & Settings \\
\hline
\endhead
\hline
\multicolumn{4}{r@{}}{Continued on next page}\\
\endfoot
\hline
\multicolumn{4}{@{}p{\textwidth}@{}}{%
\vspace{3pt}
\emph{Data input representation.} Start--stop methods use interval-level
records with time-updated covariates.  Baseline methods use one
subject-level record formed from covariates available at study entry,
with time-dependent covariate columns removed.  Landmark methods use
one subject-level record per eligible landmark risk set, using the
covariate values available at the landmark time.
}\\
\endlastfoot

RHF &
Start--stop &
\texttt{randomForestRHF} &
500 trees; tree splits $= 30$;
$\texttt{mtry}=\lceil\sqrt p\rceil$. \\[3pt]

BoXHED &
Start--stop &
\texttt{BoXHED2.0} &
Learning rate 0.1; maximum tree depth 1; 64 quantile bins; CPU
implementation with one thread; number of boosting iterations selected
by 10-fold validation. \\[3pt]

Cox-TDC &
Start--stop &
\texttt{survival::coxph} &
Efron approximation for tied event times; Breslow baseline cumulative
hazard for prediction. \\[3pt]

Baseline RSF &
Baseline &
\texttt{randomForestSRC} &
500 trees; log-rank splitting; $\texttt{mtry}=\lceil\sqrt p\rceil$;
terminal-node size 15; $\texttt{nsplit}=10$. \\[3pt]

RSF-LM &
Landmark &
\texttt{randomForestSRC} &
Dynamic multi-landmark procedure; 10 landmarks including the origin;
positive landmarks with fewer than 50 eligible training subjects were
skipped; 500 trees for each landmark RSF; otherwise the same RSF
settings as baseline RSF. \\[3pt]

LogHazNN &
Baseline &
\texttt{pycox} &
\texttt{LogisticHazard} module; two hidden layers with 32 units each;
ReLU activations; dropout 0.1; no batch normalization; Adam optimizer
with learning rate 0.001; 20 fitted time cuts; 50 epochs with
minibatches of size 256; no early stopping or simulation-specific
tuning. \\[3pt]

DeepSurv &
Baseline &
\texttt{pycox} &
\texttt{CoxPH} module; two hidden layers with 32 units each; ReLU
activations; dropout 0.1; no batch normalization; Adam optimizer with
learning rate 0.001; 50 epochs with minibatches of size 256; no early
stopping or simulation-specific tuning. \\
\end{longtable}
\endgroup

\subsection{Random hazard forests}

We applied RHF to start--stop data using the interval-level
covariates $\Z_{i,r}$.  All calculations were implemented using the
\texttt{randomForestRHF} R package \citep{randomForestRHF} with tuning
parameters as specified in Table~\ref{tab:benchmark-tuning}.
In the simulations, the supplied start--stop records are contiguous,
so the case-specific covariate path has no internal gaps.  A subject's
observed path may nevertheless end before the final benchmark time.
Since the AUC calculations require a risk marker at each evaluation
time, we use a piecewise-constant tail extension of the RHF hazard for
evaluation purposes.  The cumulative marker is computed by integrating
this evaluation hazard,
$$
\Hhat_{i,\mathrm{eval}}^{\mathrm{RHF}}(t)
=
\int_0^t
\hhat_{i,\mathrm{eval}}^{\mathrm{RHF}}(s)\,ds.
$$
The AUC markers for RHF are then taken to be
$$
M_i^{\mathrm{RHF}}(t)
=
\Hhat_{i,\mathrm{eval}}^{\mathrm{RHF}}(t),
\qquad
m_i^{\mathrm{RHF}}(t)
=
\hhat_{i,\mathrm{eval}}^{\mathrm{RHF}}(t).
$$

\subsection{Boosted nonparametric hazard estimator}

The boosted nonparametric hazard estimator of \citet{lee2021boosted}
was implemented using \texttt{BoXHED2.0} Python software
\citep{pmlr-v119-wang20o, JSSv113i03}.  BoXHED
was fit to the same start--stop training data as RHF, using the
interval start and stop times, the event indicator, the subject
identifier, and the row-level covariate vectors $\Z_{i,r}$.  The
software was run with learning rate 0.1, maximum tree depth 1, and 64
quantile bins.  The total number of boosted iterations was determined
by 10-fold validation.

Markers are obtained from the BoXHED estimated hazard.
Let
$\hhat_i^{\mathrm{BoXHED}}$ denote the case-specific predicted
hazard.  The instantaneous marker at grid value $t_k$ is
$$
m_i^{\mathrm{BoXHED}}(t_k)
=
\hhat_i^{\mathrm{BoXHED}}(t_k).
$$
The cumulative marker is obtained by accumulating this predicted hazard
over the evaluation grid,
$$
M_i^{\mathrm{BoXHED}}(t_k)
=
\Hhat_i^{\mathrm{BoXHED}}(t_k)
=
\sum_{\ell=1}^k
\hhat_i^{\mathrm{BoXHED}}(t_\ell)
(t_\ell-t_{\ell-1}).
$$

\subsection{Cox regression with time-dependent covariates}

The Cox model with time-dependent covariates is fit to the same
start--stop training data as RHF and models the hazard as
$$
\haz(t\mid \Z_i(t))=\haz_0(t)\exp\{\bbeta^T\Z_i(t)\}.
$$
The supplied covariates enter additively and linearly, with
time-constant coefficients.  For Simulation~1, these are the baseline
coordinates and the sampled time-dependent signal $w_i(t)$ from
Section~\ref{subsec:supp-sim1}; no nonlinear transformation of that
signal is added.

For discrimination at an evaluation time $t_k$, the instantaneous Cox
marker is formed from the covariate value assigned to the subject at
that time.  With $\widehat{\bbeta}$ denoting the
fitted coefficient vector, we use
$$
m_i^{\mathrm{CoxTDC}}(t_k)
=
\exp\{\widehat{\bbeta}^T\Z_i(t_k)\}.
$$
The cumulative marker accumulates the fitted Cox hazard along the same
time-updated path:
$$
M_i^{\mathrm{CoxTDC}}(t_k)
=
\int_0^{t_k}
\exp\{\widehat{\bbeta}^T\Z_i(s)\}\,
d\Hhat_0(s),
$$
where $\Hhat_0$ is the Breslow estimator of the baseline cumulative
hazard.  In computation, this integral is evaluated by summing Breslow
baseline-hazard increments with the relative-risk term evaluated from
the start--stop row active at the increment time.

\subsection{Random survival forests with baseline covariates}

Each RSF tree partitions $\RR^p$ into terminal nodes and estimates a
Nelson--Aalen cumulative hazard within each node.  The RSF forest averages
these estimates across trees \citep{ishwaran2008rsf}.  To use RSF in
the time-varying simulations, we collapse the counting-process data to
one survival row per subject and drop the simulated time-dependent
columns.  We write $\X_i$ for the resulting covariate vector.  For
subject $i$ the forest returns the cumulative hazard estimator
$\Hhat^{\mathrm{RSF}}(t \mid \X_i)$.  For the cumulative AUC we use
$$
M_i^{\mathrm{RSF}}(t) = \Hhat^{\mathrm{RSF}}(t \mid \X_i).
$$
For the incident AUC we approximate the hazard on the evaluation grid by
$$
m_i^{\mathrm{RSF}}(t_k)
=
\frac{\Hhat^{\mathrm{RSF}}(t_k \mid \X_i)-\Hhat^{\mathrm{RSF}}(t_{k-1} \mid \X_i)}
     {t_k-t_{k-1}},
\qquad k=1,\ldots,K,
$$
and treat this marker as piecewise constant on $(t_{k-1},t_k]$.  All
calculations were based on the \texttt{randomForestSRC} R package
\citep{rfsrc}, with the tuning settings in Table~\ref{tab:benchmark-tuning}.

\subsection{Landmark random survival forests}

We extend RSF using a dynamic multi-landmark approach.
Let $\mathcal L=\{\ell_0,\ldots,\ell_J\}$ denote the
landmark grid, where $\ell_0=0$ and the positive landmarks are
selected from a thinned version of the common evaluation grid $\tt$.
In the simulations, we use $J=10$ landmarks, including the origin.

At a positive landmark $\ell$, the landmark sample consists of subjects
still at risk, $Y_i(\ell)=1$, with a completed
covariate record at or before $\ell$.  For each such subject
we form the landmark covariate vector $\Z_i(\ell)$ and define the
post-landmark follow-up and event indicator by
$$
T_i^{(\ell)} = T_i-\ell,
\qquad
\delta_i^{(\ell)} = \delta_i.
$$
We fit RSF to the landmark data
$$
\{(T_i^{(\ell)},\delta_i^{(\ell)},\Z_i(\ell)):Y_i(\ell)=1\}.
$$
This yields a conditional survival estimate
$$
\widehat S^{(\ell)}(u\mid \Z_i(\ell)),\qquad u\ge 0,
$$
and the corresponding landmark-specific cumulative hazard on the
original time scale,
$$
\Hhat_i^{(\ell)}(t)
=
-\log \widehat S^{(\ell)}(t-\ell\mid \Z_i(\ell)),
\qquad t\ge \ell.
$$
To initialize the process, we fit a
baseline RSF on the static baseline covariates to supply the initial
landmark prediction path.

The dynamic landmark predictor is formed by stitching cumulative-hazard
increments from the most recent available landmark.  For the evaluation
interval $I_k=(t_{k-1},t_k]$, let
$$
\ell_{ik}
=
\max\{\ell\in\mathcal L:\ell\le t_{k-1}
\ \hbox{and subject } i \hbox{ has a prediction at } \ell\}.
$$
The interval-specific cumulative-hazard increment is
$$
\Delta\Hhat_{ik}^{\mathrm{LM}}
=
\Hhat_i^{(\ell_{ik})}(t_k)
-
\Hhat_i^{(\ell_{ik})}(t_{k-1}),
$$
where both terms are evaluated from the same landmark-specific curve.
The stitched landmark cumulative hazard and finite-difference hazard are
then
$$
\Hhat_i^{\mathrm{LM}}(t_k)
=
\sum_{r\le k}\Delta\Hhat_{ir}^{\mathrm{LM}},
\qquad
\hhat_i^{\mathrm{LM}}(t_k)
=
\frac{\Delta\Hhat_{ik}^{\mathrm{LM}}}{t_k-t_{k-1}}.
$$
The AUC markers for the dynamic landmark RSF are
$$
M_i^{\mathrm{LM}}(t_k)=\Hhat_i^{\mathrm{LM}}(t_k),
\qquad
m_i^{\mathrm{LM}}(t_k)=\hhat_i^{\mathrm{LM}}(t_k).
$$

\subsection{Discrete-time neural hazard model}

Following the logistic hazard approach for censored survival data
\citep{biganzoli1998feed,gensheimer2019scalable}, we fit a
discrete-time neural hazard model using the \texttt{LogisticHazard}
module from \texttt{pycox}
\citep{kvamme2019time,kvamme2021continuous}.  The model was fit to the
same subject-level baseline survival representation used by RSF, with
time-dependent covariate columns removed before fitting.  The fitting
time scale was discretized from the training durations using 20 cut
points.  The fitted survival curves were then aligned to the common
benchmark grid $\tt$ before constructing the AUC markers and the hazard
accuracy metric.

For subject $i$ with baseline covariates $\X_i$, the discrete-time
hazard model represents conditional event probabilities over the fitted
time intervals.  These probabilities are modeled using a logistic link,
$$
\mathrm{logit}\{p_{ik}\}
=
f_\theta(\X_i,k),
$$
where $f_\theta$ is a fully connected neural network with parameters
$\theta$.  The fitted discrete survival curve is converted to the
benchmark grid and the cumulative marker is
$$
M_i^{\mathrm{NH}}(t_k)
=
-\log \widehat{S}_i(t_k),
\qquad k = 1,\ldots,K.
$$
For the incident AUC we use an instantaneous marker based on increments,
$$
m_i^{\mathrm{NH}}(t_k)
=
\frac{M_i^{\mathrm{NH}}(t_k) - M_i^{\mathrm{NH}}(t_{k-1})}{t_k - t_{k-1}},
\qquad k = 1,\ldots,K,
$$
with $M_i^{\mathrm{NH}}(t_0):=0$.

The neural network used two hidden layers with 32 units per layer,
ReLU activations, dropout probability 0.1, and no batch
normalization.  Models were fit with the Adam optimizer using learning
rate 0.001 for 50 epochs with minibatches of size 256.  No early
stopping or simulation-specific hyperparameter tuning was used.

\subsection{Deep neural Cox model}

Let $\X_i$ denote the subject-level covariate vector used for the
baseline neural methods, and let $T_i$ be the event time with
indicator $\delta_i$.  The hazard is modeled as
$$
\haz(t \mid \X_i)
=
\haz_0(t)\,\exp\{ f_\theta(\X_i) \},
$$
where $\haz_0(t)$ is an unspecified baseline hazard and
$f_\theta(\X_i)$ is the output of a fully connected neural network
\citep{faraggi1995neural,katzman2018deepsurv}.  Parameters $\theta$
are estimated by maximizing the Cox partial likelihood, and a
Breslow-type estimator $\Hhat_0(t)$ is used for the baseline
cumulative hazard.

For subject $i$ this yields the cumulative marker
$$
M_i^{\mathrm{DS}}(t)
=
\Hhat_0(t)\,\exp\{\widehat{f}_\theta(\X_i)\}
=
\Hhat(t \mid \X_i).
$$
For the incident AUC we use increments of this estimated cumulative
hazard on the evaluation grid,
$$
m_i^{\mathrm{DS}}(t_k)
=
\frac{M_i^{\mathrm{DS}}(t_k)-M_i^{\mathrm{DS}}(t_{k-1})}
     {t_k-t_{k-1}},
\qquad k=1,\ldots,K.
$$
When the fitted baseline hazard has a positive increment over the grid
cell, this marker is proportional to
$\exp\{\widehat{f}_\theta(\X_i)\}$ and therefore preserves the Cox
ranking at that time.  We implement this model using the
\texttt{CoxPH} module from \texttt{pycox}
\citep{kvamme2019time,kvamme2021continuous}.  The same neural-network
architecture and training settings used for the discrete-time neural
hazard model were used for DeepSurv: two hidden layers with 32 units
per layer, ReLU activations, dropout probability 0.1, and no batch
normalization.  Models were fit with the Adam optimizer using learning
rate 0.001 for 50 epochs with minibatches of size 256.  No early
stopping or simulation-specific hyperparameter tuning was used.

\section{Estimation of relative hazard mean squared error}
\label{sec:supp-hazard-metrics}

The relative hazard mean squared error is defined in Section~5.2 of the
main paper.  This section gives the empirical estimator used in the
simulation studies.  For a test sample of size $n$, the reported empirical
relative hazard mean squared error is
$$
\frac{
n^{-1}\sum_{i=1}^n\sum_{k=1}^{K}
\{\hhat(t_k,\X_i(t_k))-\haz(t_k,\X_i(t_k))\}^2
}{
n^{-1}\sum_{i=1}^n\sum_{k=1}^{K} 
\{\haz(t_k,\X_i(t_k))\}^2
}.
$$
Here $\hhat(t_k,\X_i(t_k))$ denotes the method-specific hazard estimate used for
subject $i$ at evaluation time $t_k$.

\section{Benchmark results with a smaller sample size}
\label{subsec:supp-benchmark-n500}

\begin{figure}[phtb]
\centering
\includegraphics[width=\textwidth]{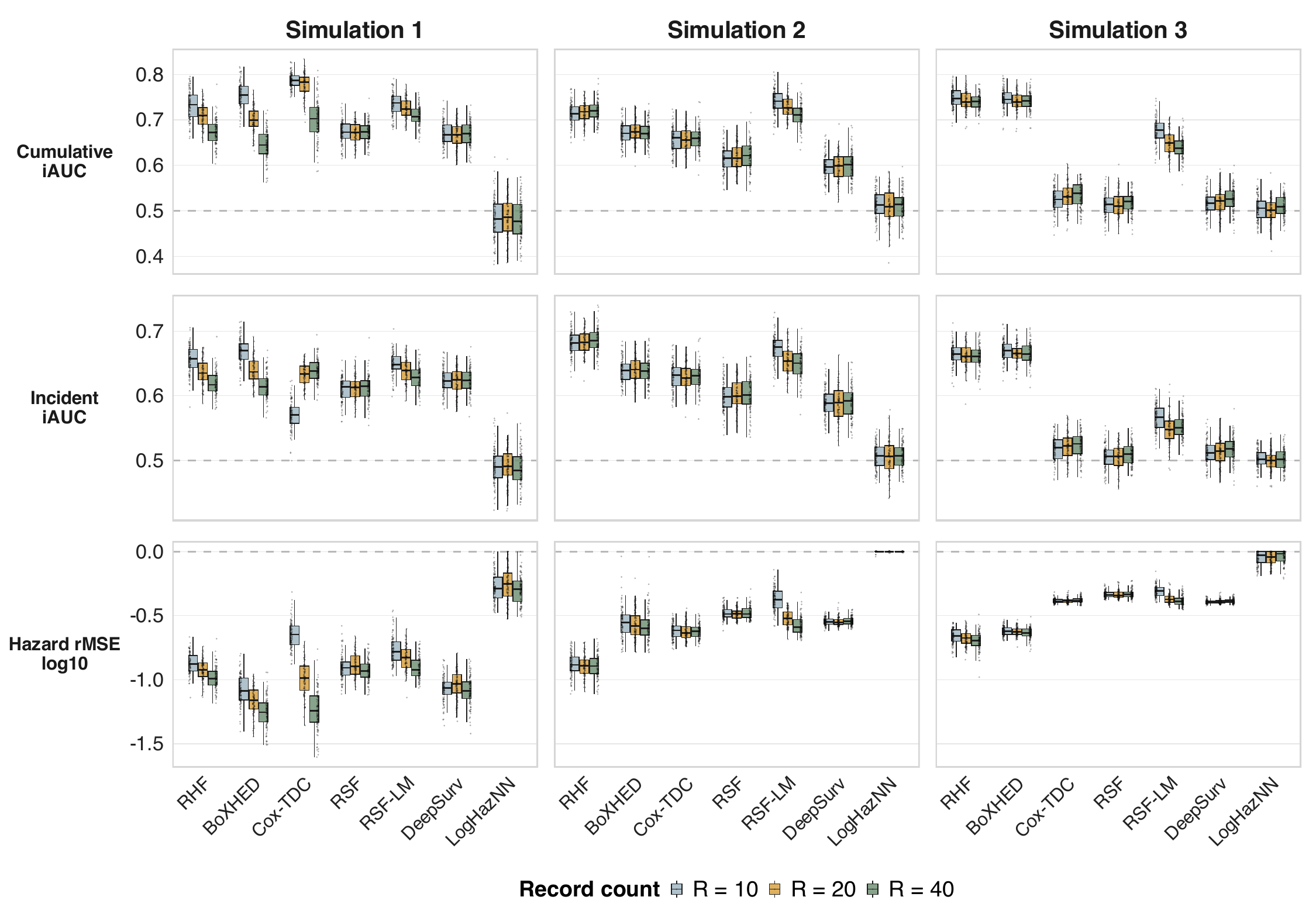}
\vskip-10pt
\caption{Benchmark performance with independent training and test
samples of size $n=500$ in the three time-varying covariate
simulations. Boxplots and overlaid points show replicate-level
results grouped by nominal record count $R=10,20,40$. Higher values
indicate better discrimination for cumulative and incident iAUC;
lower values indicate more accurate hazard estimation for hazard
rMSE. Hazard rMSE is displayed on the $\log_{10}$ scale, with extreme
values clipped for readability before constructing the boxplots.
Dashed reference lines denote $\mathrm{iAUC}=0.5$ and
$\log_{10}(\mathrm{rMSE})=0$.}
\label{fig:supp-benchmark-n500}
\end{figure}

We repeated the three simulation experiments of the main paper with
training and test samples of size $n=500$ to assess whether the
conclusions of the main paper hold at a smaller sample size.  The
data-generating mechanisms, nominal record counts $R=10,20,40$,
tuning settings (Table~\ref{tab:benchmark-tuning}), and evaluation
conventions are unchanged from the $n=2500$ experiments, and each
simulation and record-count setting uses 100 independent Monte Carlo
replicates.  Figure~\ref{fig:supp-benchmark-n500} displays cumulative
and incident iAUC and hazard rMSE for the seven methods.

The results are generally consistent with the main experiments, which
used $n=2500$.  RHF has the lowest hazard rMSE in Simulations~2 and~3
and remains among the top methods for discrimination.  In
Simulation~3, RHF and BoXHED are clearly separated from the remaining
methods on both discrimination measures.  In the smoother
Simulation~1, Cox-TDC continues to provide high cumulative
discrimination, while its hazard rMSE decreases as $R$ increases.

\section{Missingness in the MIMIC-IV predictors}
\label{sec:supp-mimic-missingness}

Table~\ref{tab:mimic-missingness} summarizes missingness for 62 MIMIC-IV
predictors after hourly feature construction, before the subsequent
6-hour carry-forward for vital signs and respiratory settings,
24-hour carry-forward for laboratory values, and fixed-value
imputation. For each predictor, the count is the number of hourly
start--stop records with a missing value at this stage. Percentages
use a common denominator of $13{,}869{,}748$ hourly records, with each
record contributing equally.

Hourly feature construction had already incorporated admission-time
lookback values and predictor-specific histories. In particular, the
latest Glasgow Coma Scale (GCS) predictor carried forward the most
recent available GCS value within each stay, whereas its six-hour
mean used available hourly GCS values within a full six-bin window.
Current treatment-use indicators were coded as zero or one. Their
six-hour summaries required a full six-bin history and were missing
in the initial records before that history was available.

At this stage, the latest GCS was missing in $1.24\%$ of records,
compared with $65.07\%$ for its six-hour mean. Among the laboratory
predictors, missingness was $94.47\%$ for blood urea nitrogen,
$94.54\%$ for bicarbonate, $94.73\%$ for white blood cell count,
$94.89\%$ for platelet count, and $98.08\%$ for lactate. Heart rate
and oxygen saturation had missingness of $62.45\%$ and $63.14\%$,
respectively. Age and the current treatment-use indicators had no
missing entries; each corresponding six-hour treatment summary had
$2.35\%$ missingness. These percentages describe missingness before
the subsequent gap-filling steps, rather than the percentage of
records requiring fixed-value imputation.

\begingroup
\footnotesize
\setlength{\tabcolsep}{5pt}
\renewcommand{\arraystretch}{1.12}
\setlength{\LTcapwidth}{\textwidth}
\begin{longtable}{@{}p{0.62\textwidth}rr@{}}
\caption{Missingness in MIMIC-IV predictors after hourly feature
construction, before the subsequent 6-hour and 24-hour carry-forward
and fixed-value imputation. Predictor-specific histories, including
the carried-forward latest GCS, had already been constructed.
Counts refer to hourly start--stop records with a missing predictor
value at this stage. Percentages use $13{,}869{,}748$ records as the
common denominator and are rounded to two decimal places.}
\label{tab:mimic-missingness}\\
\hline
Predictor & Missing records & Missing (\%) \\
\hline
\endfirsthead
\multicolumn{3}{@{}l}{Table~\thetable\ (continued)}\\[3pt]
\hline
Predictor & Missing records & Missing (\%) \\
\hline
\endhead
\hline
\multicolumn{3}{r@{}}{Continued on next page}\\
\endfoot
\hline
\endlastfoot
Age (years) & 0 & 0.00 \\
Heart rate (beats/min) & 8,661,918 & 62.45 \\
Systolic blood pressure (mmHg) & 9,078,712 & 65.46 \\
Diastolic blood pressure (mmHg) & 9,079,231 & 65.46 \\
Mean arterial pressure (mmHg) & 10,677,839 & 76.99 \\
Respiratory rate (breaths/min) & 8,726,858 & 62.92 \\
Oxygen saturation (SpO$_2$) (\%) & 8,757,934 & 63.14 \\
Temperature ($^\circ$C) & 12,189,394 & 87.88 \\
Supplemental oxygen in use (0/1) & 0 & 0.00 \\
Supplemental oxygen in past 6h & 325,874 & 2.35 \\
Glasgow Coma Scale (GCS) & 171,986 & 1.24 \\
GCS mean, past 6h (score) & 9,024,520 & 65.07 \\
Creatinine (mg/dL) & 13,091,077 & 94.39 \\
Blood urea nitrogen (BUN) (mg/dL) & 13,102,617 & 94.47 \\
Sodium (mEq/L) & 13,008,116 & 93.79 \\
Potassium (mEq/L) & 12,951,097 & 93.38 \\
Chloride (mEq/L) & 13,047,938 & 94.07 \\
Bicarbonate (mEq/L) & 13,111,768 & 94.54 \\
Calcium (mg/dL) & 13,044,310 & 94.05 \\
Magnesium (mg/dL) & 13,143,584 & 94.76 \\
Phosphate (mg/dL) & 13,202,112 & 95.19 \\
Glucose (mg/dL) & 12,993,596 & 93.68 \\
Aspartate aminotransferase (AST) (U/L) & 13,657,574 & 98.47 \\
Alanine aminotransferase (ALT) (U/L) & 13,659,827 & 98.49 \\
Alkaline phosphatase (U/L) & 13,661,947 & 98.50 \\
Albumin (g/dL) & 13,761,631 & 99.22 \\
Total bilirubin (mg/dL) & 13,659,925 & 98.49 \\
White blood cell count ($10^3/\mu\mathrm{L}$) & 13,138,338 & 94.73 \\
Hemoglobin (g/dL) & 13,116,818 & 94.57 \\
Hematocrit (\%) & 13,063,167 & 94.18 \\
Platelet count ($10^3/\mu\mathrm{L}$) & 13,161,621 & 94.89 \\
INR & 13,450,499 & 96.98 \\
Prothrombin time (PT) (s) & 13,450,386 & 96.98 \\
Activated partial thromboplastin time (aPTT) (s) & 13,417,737 & 96.74 \\
Troponin & 13,822,665 & 99.66 \\
Creatine kinase (CK) (U/L) & 13,791,377 & 99.43 \\
C-reactive protein (CRP) & 13,859,835 & 99.93 \\
PaCO$_2$ (mmHg) & 13,453,446 & 97.00 \\
Arterial pH & 13,360,549 & 96.33 \\
Lactate (mmol/L) & 13,603,892 & 98.08 \\
Urine output (mL per bin) & 11,361,070 & 81.91 \\
Urine output, past 6h (mL) & 12,772,348 & 92.09 \\
Norepinephrine infusion in use (0/1) & 0 & 0.00 \\
Norepinephrine in past 6h & 325,874 & 2.35 \\
Epinephrine infusion in use (0/1) & 0 & 0.00 \\
Epinephrine in past 6h & 325,874 & 2.35 \\
Dopamine infusion in use (0/1) & 0 & 0.00 \\
Dopamine in past 6h & 325,874 & 2.35 \\
Dobutamine infusion in use (0/1) & 0 & 0.00 \\
Dobutamine in past 6h & 325,874 & 2.35 \\
Vasopressin infusion in use (0/1) & 0 & 0.00 \\
Vasopressin in past 6h & 325,874 & 2.35 \\
Phenylephrine infusion in use (0/1) & 0 & 0.00 \\
Phenylephrine in past 6h & 325,874 & 2.35 \\
Mechanical ventilation in use (0/1) & 0 & 0.00 \\
Mechanical ventilation in past 6h & 325,874 & 2.35 \\
Fraction of inspired oxygen (FiO$_2$) & 12,768,302 & 92.06 \\
Positive end-expiratory pressure (PEEP) (cmH$_2$O) & 13,300,813 & 95.90 \\
Arterial oxygen partial pressure (PaO$_2$) (mmHg) & 13,453,174 & 97.00 \\
PaO$_2$/FiO$_2$ ratio (mmHg) & 13,756,881 & 99.19 \\
Renal replacement therapy (RRT) in use (0/1) & 0 & 0.00 \\
RRT in past 6h & 325,874 & 2.35 \\
\end{longtable}
\endgroup

\section{Sensitivity to including missingness indicators}
\label{sec:supp-mimic-indicator-sensitivity}

The primary MIMIC-IV analysis excluded missingness indicators. We
repeated the RHF and time-localized variable-priority analyses with
62 pre-imputation missingness indicators added to the 68 original
predictors. Each indicator records whether the corresponding value
was missing after hourly feature construction, before the subsequent
carry-forward and fixed-value imputation steps, as described in
Section~\ref{sec:supp-mimic-missingness}. 

\begin{figure}[htbp]
\centering
\includegraphics[width=\textwidth]{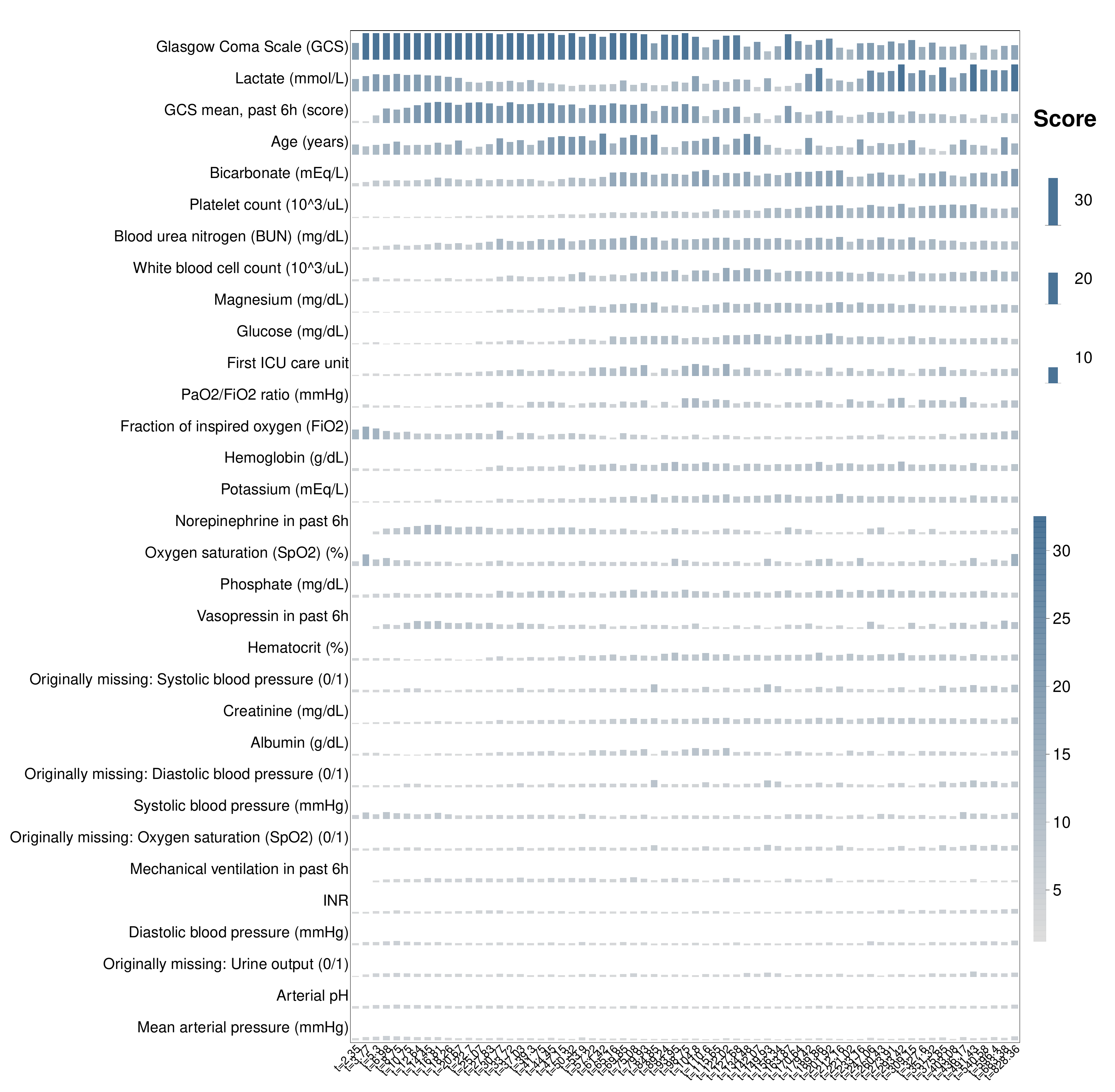}
\caption{Time-localized MIMIC-IV variable priority with pre-imputation
missingness indicators included. The 32 displayed variables appeared
among the 15 largest finite scores in at least one of the 65
evaluation windows. Rows are ordered by the 90th percentile of the
finite window-specific scores. Each column represents an evaluation
window and is labeled by its right endpoint in hours since ICU
admission; columns are equally spaced despite unequal window
lengths. Bar height and color encode priority. For visualization,
heights and colors are capped at the 99th percentile of the plotted
scores (32.51); Table~\ref{tab:mimic-indicator-sensitivity} uses the
original, uncapped values. Labels beginning ``Originally missing''
identify the added indicators.}
\label{fig:mimic-indicator-sensitivity}
\end{figure}

Figure~\ref{fig:mimic-indicator-sensitivity} shows the time-localized
priority profiles, and Table~\ref{tab:mimic-indicator-sensitivity}
summarizes the original scores.  Neurologic status and lactate had
the largest scores.  The latest Glasgow Coma Scale (GCS) had large
scores throughout follow-up, its six-hour mean had especially large
scores early, and lactate had larger scores later in follow-up.  Age,
bicarbonate, and blood urea nitrogen also had substantial priority,
while platelet count and several other laboratory predictors had
larger scores later.

Missingness indicators had lower overall priority than the leading
clinical predictors. Ranking variables by the 90th percentile of
their finite window-specific scores, the highest-ranked indicator
was systolic blood pressure missingness (rank 21; score 7.40),
followed by diastolic blood pressure, oxygen saturation, and urine
output missingness (ranks 24, 26, and 34). For comparison, the scores
for the latest GCS, lactate, and six-hour mean GCS were 37.75, 25.72,
and 24.18, respectively. The missingness indicators for lactate,
bicarbonate, blood urea nitrogen, platelet count, and white blood
cell count had zero reported priority at all 65 evaluation windows.
Thus, some indicators contributed to the fitted priority profiles,
but the leading scores were assigned to clinical predictors.

\begin{table}[htbp]
\centering
\caption{Variable-priority summaries with pre-imputation missingness
indicators included. The table contains the ten highest-ranked
predictors and the four missingness indicators appearing in
Figure~\ref{fig:mimic-indicator-sensitivity}. Ranks are based on the
90th percentile of finite window-specific scores among the 112
variables with at least one finite score. Every variable listed
here has finite scores in all 65 windows. Summaries use the original,
uncapped scores and weight evaluation windows equally.}
\label{tab:mimic-indicator-sensitivity}
\begingroup
\footnotesize
\setlength{\tabcolsep}{5pt}
\renewcommand{\arraystretch}{1.12}
\begin{tabular}{@{}rp{0.52\textwidth}rr@{}}
\hline
Rank & Predictor & 90th percentile & Maximum \\
\hline
\multicolumn{4}{@{}l}{\emph{Leading clinical predictors}}\\[2pt]
1 & Glasgow Coma Scale (GCS) & 37.75 & 42.88 \\
2 & Lactate & 25.72 & 39.63 \\
3 & GCS mean, past 6h & 24.18 & 25.43 \\
4 & Age & 21.88 & 25.71 \\
5 & Bicarbonate & 17.92 & 20.95 \\
6 & Platelet count & 15.23 & 17.22 \\
7 & Blood urea nitrogen & 14.51 & 16.72 \\
8 & White blood cell count & 13.25 & 16.14 \\
9 & Magnesium & 12.27 & 13.13 \\
10 & Glucose & 10.96 & 13.18 \\
\hline
\multicolumn{4}{@{}l}{\emph{Missingness indicators}}\\[2pt]
21 & Systolic blood pressure: missing & 7.40 & 9.84 \\
24 & Diastolic blood pressure: missing & 6.72 & 8.77 \\
26 & Oxygen saturation (SpO$_2$): missing & 5.95 & 6.90 \\
34 & Urine output: missing & 4.53 & 6.56 \\
\hline
\end{tabular}
\endgroup
\end{table}

\clearpage

\bibliographystyle{imsart-nameyear}
\bibliography{references}